\documentclass{article}
\usepackage{titlesec}
\usepackage{subcaption}
\usepackage{makecell} 
\usepackage[utf8]{inputenc}
\usepackage{float}
\usepackage{amssymb}
\usepackage{hyperref, array, supertabular, tabulary}
\usepackage{longtable}
\usepackage{algorithm}
\usepackage{algpseudocode}
\usepackage{times} 
\usepackage[colorinlistoftodos,color=pink]{todonotes} 
\usepackage{xparse}
\usepackage[bottom=0.65in,top=0.7in,left=0.7in,right=0.7in]{geometry}
\usepackage{multirow}
\usepackage{url}
\usepackage{setspace}
\algnewcommand\algorithmicforeach{\textbf{for each}}
\algdef{S}[FOR]{ForEach}[1]{\algorithmicforeach\ #1\ \algorithmicdo}
\titleformat*{\section}{\normalsize\bfseries}
\titleformat*{\subsection}{\normalsize\itshape}
\titleformat*{\subsubsection}{\normalsize\itshape}
\begin{document}

\title{\vspace{-2.0cm}The UU-test for Statistical Modeling of Unimodal Data}

\author{Paraskevi Chasani and Aristidis Likas$$ \thanks{Corresponding author}\\ \normalsize Department of Computer Science and Engineering \vspace{-0.4cm}\\ \normalsize University of Ioannina \vspace{-0.4cm}\\ 
	\normalsize GR 45110, Ioannina, Greece \vspace{-0.4cm}\\
	\normalsize e-mail: \{pchasani, arly\}@cs.uoi.gr}

\date{}
\maketitle
\vspace{-0.5cm}
\par\noindent\rule{\textwidth}{0.4pt}

\textbf{Abstract} 

Deciding on the unimodality of a dataset is an important problem in data analysis and statistical modeling. It allows to obtain knowledge about the structure of the dataset, i.e. whether data points have been generated by a probability distribution with a single or more than one peaks. Such knowledge is very useful for several data analysis problems, such as for deciding on the number of clusters and determining unimodal projections. 
We propose a technique called UU-test (Unimodal Uniform test) to decide on the unimodality of a one-dimensional dataset. The method operates on the empirical cumulative density function (ecdf) of the dataset. It attempts to build a piecewise linear approximation of the ecdf that is unimodal and models the data sufficiently in the sense that the data corresponding to each linear segment follows the uniform distribution. A unique feature of this approach is that in the case of unimodality, it also provides a statistical model of the data in the form of a Uniform Mixture Model. We present experimental results in order to assess the ability of the method to decide on unimodality and perform comparisons with the well-known dip-test approach. In addition, in the case of unimodal datasets we evaluate the Uniform Mixture Models provided by the proposed method using the test set log-likelihood and the two-sample Kolmogorov-Smirnov (KS) test.

\hspace{-1.3em}\emph{Keywords}: unimodal data, unimodality test, statistical modeling, uniform mixture model
\par\noindent\rule{\textwidth}{0.4pt}

\vspace{-0.5cm}
\section{Introduction}
Gaining knowledge of data distributions is a significant topic in data analysis. Many problems require assumptions related to the shape of the data in order to be solved. It is of considerable importance to understand the grouping behavior of points, i.e. whether the data are 'gathered' or not. Data points generated by \textit{unimodal distributions}, such as Gaussian, Student’s t, Gamma etc., exhibit such a grouping behavior and form single and coherent clusters. Unimodal distributions constitute a wide family, and methods or tests to determine unimodality of a dataset are of particular significance. Although a great deal of research work focused on Gaussianity (or normality) tests, few methods have been proposed for the more general problem of deciding distribution unimodality. In this work, given a 1-d dataset, we propose a technique (called UU-test) that builds a unimodal statistical model in order to fit unimodal data. An appealing feature of the method is that it performs statistical data modeling and decides on distribution unimodality, simultaneously. 

In order to decide whether a set of data points has been generated by a probability distribution that is unimodal or not (multimodal), we use unimodality tests. The importance of such a test is high, since it provides fundamental information about the data structure. For example, it is reasonable to apply clustering methods, only when cluster structure is present in the dataset. In \cite{adolfsson2019cluster} it is presented in a nice way why a unimodality test is important in cluster analysis. A test for the existence of cluster structure should precede the application of clustering algorithms, as the success of any clustering algorithm depends on the existence of underlying cluster structure. Moreover, in incremental clustering methods, a unimodality test could be employed to determine whether to split (or not) an already found cluster \cite{kalogeratos2012dip}. 

Several tests have been proposed to decide on unimodality/multimodality of a dataset. The most widely used is \textit{Hartigans' dip-test} \cite{hartigan1985dip} that computes the \textit{dip statistic}, a distance between the data empirical cdf from the closest unimodal cdf. Note that the dip-test typically applies to 1-d data. A more recently proposed general unimodality test is the \textit{folding test} \cite{siffer2018your}, which can be used for multidimensional data. 

In this work we propose the \textit{UU-test} method for modeling one dimensional data generated by unimodal distributions. It works with the empirical cumulative density function (ecdf) of the data, assuming the data distribution is continuous. It is important to note that UU-test does not make use of any parameters. In addition, it relies on well-known uniformity tests (e.g. Kolmogorov-Smirnov \cite{dodge2008kolmogorov}), thus it does not require the computation of bootstrap samples (like Hartigans’ dip-test), a fact that saves computational time. Note also that all other tests focus on the decision on distribution unimodality and do not address the problem of statistical modeling of unimodal data. On the contrary our approach, in the case of unimodality,  provides also a statistical model of the data in the form of a \textit{Uniform Mixture Model} (UMM).

The outline of the paper is as follows: in Section \ref{Relatedwork}, we present related work on unimodality tests and describe their importance on data analysis tasks. In Section \ref{Def} we provide the necessary definitions and notations. Then we present the proposed UU-test method in Section \ref{UU} and attempt to explain the method using several illustrative examples. In Section \ref{Modeling}, we present the statistical model of unimodal data in the form of a uniform mixture model provided by UU-test. Experimental results are provided in Section \ref{Experiments} aiming at evaluating both the decisions of the method as well as the performance of the constructed uniform mixture model. Section \ref{Multipledim} refers to unimodality in multiple dimensions while appropriate cut points are suggested by UU-test in order to split multimodal datasets into unimodal subsets. Finally, in Section \ref{Conclusions} we provide conclusions and directions for future work.

\section{Related Work} \label{Relatedwork}

Unimodality tests are used to decide whether a set of data points has been generated by a probability distribution with a single mode (peak). The unimodality property is directly related to the grouping behavior of points, i.e. whether data are ‘gathered’ or not. Detecting unimodality is very useful in several data analysis applications, e.g. clustering, feature selection etc. The most typical example of unimodality is normality (or Gaussianity), which can be tested using several well-known tests, for example the Anderson-Darling test \cite{anderson1952} and Shapiro-Wilk test \cite{10.1093/biomet/52.3-4.591}. For this reason, in several data analysis methods, the normality test has been used to check the grouping behavior of data. It is obvious that the employment of normality tests to check unimodality relies on a crude assumption, since there are many datasets whose density (e.g. histogram) has a single peak (i.e. they are unimodal) but its shape does not resemble the shape of the normal distribution. It is obvious that in such cases a normality test will fail. Therefore, the development of general unimodality tests offers great advantage compared to normality tests, since it allows to test the ‘gathering property’ of data without focusing on a particular functional form (e.g. Gaussian, Student’s t, uniform, Gamma, truncated Gaussian etc). 

Several unimodality tests have been proposed in the literature, most of them applied to 1-d data. The oldest one is Silverman's test \cite{silverman1981using}, also known as \textit{bandwidth test}. It applies kernel density estimation with Gaussian kernel and relies on the kernel bandwidth to decide on unimodality. Note that kernel bandwidth is related to the amount of smoothing. If high bandwidth (i.e. large smoothing) is needed to obtain a unimodal
estimate, this in an indication of multimodality. The above idea is well-studied and several weaknesses have been identified \cite{hall2001calibration}.

Hartigans’ \textit{dip-test} \cite{hartigan1985dip} constitutes the most popular unimodality test. Given a 1-d dataset, it computes the \textit{dip statistic} as the maximum difference between the ecdf of the data and the unimodal distribution function that minimizes that maximum difference. The uniform distribution is the asymptotically least favorable unimodal distribution, and the distribution of the test statistic (dip statistic) is determined asymptotically and empirically through sampling from the uniform distribution. Given a set of real numbers \(X=\{x_1,x_2,...,x_n\}\) the dip-test computes the $dip(X)$ value, which is the departure from unimodality of the ecdf. In other words, the dip statistic computes the minimum among the maximum deviations observed between the ecdf $F$ and the cdfs from the family of unimodal distributions. The dip-test returns not only the dip value, but also the statistical significance of the computed dip value, i.e. a $p$-value. To compute the $p$-value, the class of uniform distributions \textit{U} is used as the null hypothesis, since its dip values are stochastically larger than other unimodal distributions, such as those having exponentially decreasing tails. The computation of the $p$-value uses $b$ bootstrap sets $U_n^r$ ($r=1,...,b$) of $n$ observations each sampled from the $U[0,1]$ uniform distribution. $P$-value is computed as the probability of $dip(X)$ being less than the $dip(U_n^r)$:
\begin{displaymath}\ P=\#[dip(X)\leq{dip(\textit{$U_n^r$})}]/b
\end{displaymath}

Another 1-d unimodality test is the \textit{excess mass test} \cite{10.2307/2290406} that measures the excess mass of the modes, i.e. the amount of density (as estimated by a histogram) that is above a specific level L. If this excess mass is distributed in several regions, then this is an indication of multimodality.

The \textit{RUNT test} \cite{hartigan1992runt} and the \textit{MAP test} \cite{rozal1994map} constitute attempts to address the unimodality issue in multiple dimensions. RUNT test is based on single linkage clustering, while MAP test uses minimum trees with additional constraints, thus both approaches are computationally expensive. 

In \cite{siffer2018your}, the \textit{folding test} is presented, which is simple test that can be applied in both univariate and multivariate cases. It relies on the folding idea that works as follows: (1) fold up the distribution with respect to a pivot $s^{\star}$, (2) compute the variance of the folded distribution and (3) compare the folded variance with the initial variance. The main idea is that the density of the folded distribution is expected to have a far lower variance in the case of multimodal distributions compared to unimodal cases. Successful application of the method requires the identification of the correct pivot, i.e. the one which provides significant variance reduction through the folding process (if such a pivot exists). A simple heuristic approach is suggested for pivot computation. 

In what concerns the use of unimodality tests in data analysis, although the dip-test applies on 1-d data, it has been mainly used in the case of multidimensional datasets. In such cases, dataset unimodality can be estimated by performing several 1-d tests. For example, in \cite{kalogeratos2012dip} the \textit{dip-dist} criterion has been proposed to determine whether a data subset is unimodal or not. It relies on the application of dip-test on each row of the pairwise distance matrix of the data. This criterion has been integrated into a clustering method (called dip-means \cite{kalogeratos2012dip}) which is an incremental algorithm based on cluster splitting that uses the dip-dist to decide whether to split a cluster or not. In this way, the method is able to automatically determine the number of clusters. Another method which uses dip-test for multidimensional data clustering is SkinnyDip \cite{maurus2016skinny}. For univariate data clustering, a method called \textit{UniDip} is proposed that exploits dip-test to “pick off” one mode at a time from the data sample. For multidimensional data clustering, they wrap UniDip with a recursive heuristic over the dimensions of the data space, in order to extract the modal hyperintervals from a continuous multivariate distribution. Also the folding test has been proposed as a tool that could help clustering algorithms, such as DBSCAN \cite{siffer2018your}.

Dip-test has also been used in the DipTransformation method \cite{schelling2018diptransformation,schelling2020dataset} to improve the structure of a dataset and achieve better clustering results using k-means. Moreover, in \cite{krause2005multimodal} it is demonstrated how the dip-test can be used in projection pursuit to discover information rich low dimensional linear projections of high-dimensional data.

The proposed UU-test approach exhibits analogy to the dip-test methodology, i.e. it is applied on 1-d datasets and works with the ecdf of the dataset. However, instead of computing the distance of the ecdf from the family of unimodal distributions (dip-test), it attempts to define a unimodal distribution whose cdf sufficiently approximates the ecdf, i.e. the obtained distribution is both unimodal and a good statistical model of the dataset. In this way, in the case where unimodality is detected, we also obtain a generative model of the dataset in the form of a mixture of uniform distributions. Therefore, the method has a clear advantage over the dip-test.

\section{Notation and Definitions} \label{Def}
At first, we provide the main definitions needed to present and clarify our method. Next, we explain UU-test methodology providing Algorithms and illustrative Figures. Finally, we present the uniform mixture model that is directly provided by the method in the case where the dataset is characterized as unimodal.

Let $X=\{x_1,...,x_N\}$, $x_i \in \mathbb{R}$ and $x_i < x_{i+1}$ an ordered 1-d dataset of distinct real numbers. 
For an interval [a,b], we define $X(a,b)=\{a \leq x_i \leq b, x_i\in X\}$ the subset of $X$ whose elements belong to that interval. 
Moreover, we denote as $F_X (x)$ the empirical cumulative distribution function (ecdf) of $X$, defined as:

\begin{displaymath}
F_X(x) = \frac{\textup{number\, of\, elements\, in\, the\, sample} \leq x}{N} =
\frac{1}{N} \sum_{i=1}^N I_{(-\infty,x)}(x_i)
\end{displaymath} 
$I_{(-\infty, x)}(x_i)$ is the indicator function:
$I_{(-\infty,x)}(x_i)=\left\{
\begin{array}{ll}
      1, & \textup{if} \,\, x_i\leq x \\
      0, & \textup{otherwise}\\
\end{array} 
\right. $ 
It also holds that $F_X(x)=0$ if $x < x_1$, $F_X(x)=1$ if $x\geq x_N$. Note also that $F_X(x)$ is \textit{piecewise constant}.

In what concerns the unimodality of a distribution there are two definition options. The first relies on the probability density function (pdf): a pdf is unimodal, if it has a single mode; a region where the density becomes maximum, while non-increasing density is observed when moving away from the mode. In other words, a pdf $f(x)$ is a unimodal function if for some value $m$, it is monotonically increasing for $x\leq m$ and monotonically decreasing for $x\geq m$. In that case, the maximum value of $f(x)$ is $f(m)$ and there are no other local maxima. The second definition option relies on the cumulative distribution function (cdf): a cdf $F(x)$ is unimodal if there exist two points $x_l$ and $x_u$ such that $F(x)$ can be divided into three parts: a) a convex part $(-\infty,x_l)$, b) a constant part $[x_l,x_u]$ and c) a concave part $(x_u,\infty)$. It is worth mentioning that it is possible for either the first two parts or the last two parts to be missing. It should be stressed that the uniform distribution is unimodal and its cdf is linear.  

\begin{figure}[H]  
	\centering
    \begin{subfigure}[b]{0.4\linewidth}
    \includegraphics[width=\linewidth]{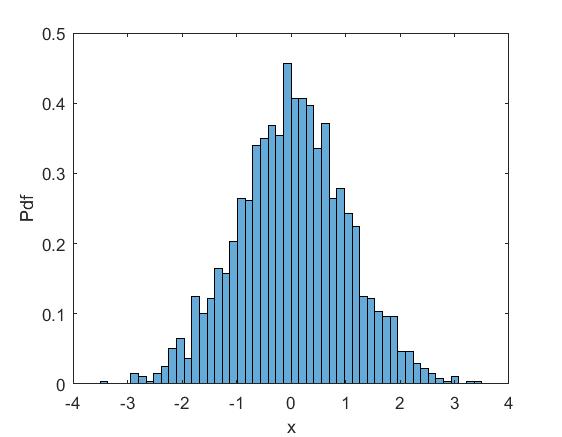}
    \caption{}
    \label{figure:fig1a}
  \end{subfigure}
 \begin{subfigure}[b]{0.4\linewidth}
    \includegraphics[width=\linewidth]{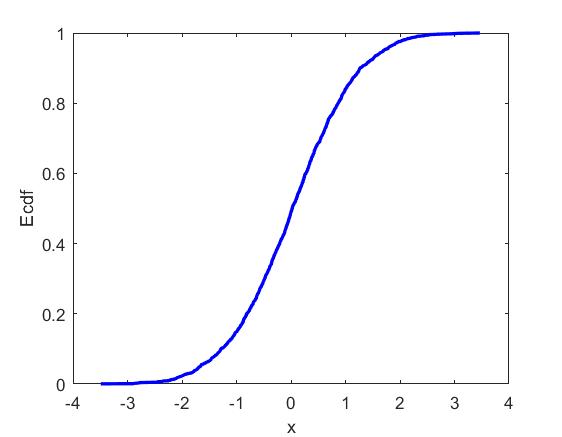}
    \caption{}
    \label{figure:fig1b}
  \end{subfigure}   
    \begin{subfigure}[b]{0.4\linewidth}
    \includegraphics[width=\linewidth]{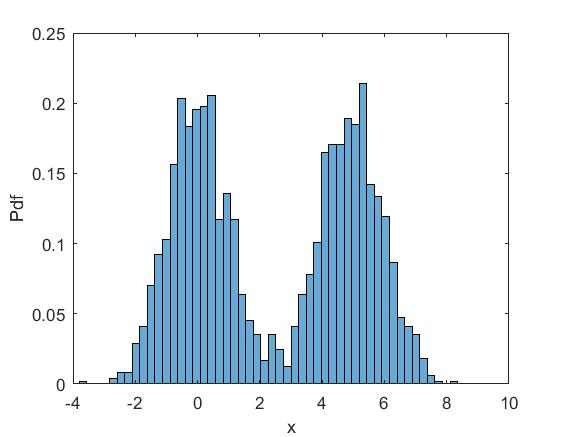}
    \caption{}
    \label{figure:fig1c}
    \end{subfigure}
    \begin{subfigure}[b]{0.4\linewidth}
    \includegraphics[width=\linewidth]{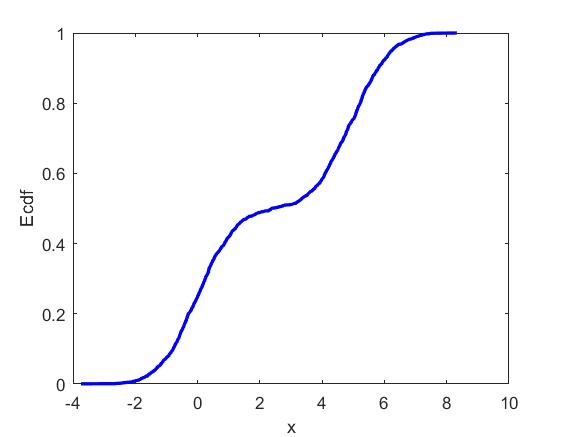}
    \caption{}
    \label{figure:fig1d}
  \end{subfigure}
  \caption{Examples of unimodal (top row) and bimodal (bottom row) datasets. In (a) and (c) the histograms are presented, while in (b) and (d) the corresponding ecdfs.}
\label{figure:fig1}
  \end{figure}

A distribution that is not unimodal is called multimodal with two or more modes. Those modes typically appear as distinct peaks (local maxima) in the pdf plot. A distribution with exactly two modes is called bimodal. Fig.~\ref{figure:fig1} illustrates typical examples of unimodal and bimodal datasets in terms of pdf plots (histograms) and ecdf plots.

It is clear from Fig.~\ref{figure:fig1} that understanding data unimodality when observing pdf plots (histograms) is simple, while in the case of ecdf plots is less straightforward. However, the shape of a histogram (used to visualize the dataset pdf) varies depending on the selected number of bins (buckets), while the ecdf plot is independent of any parameter. So, the ecdf has clear advantage over the histograms and it is used in our method.

Let a subset $S=\{s_1,...,s_L\}$ of $X (s_i \in X)$ with $s_i \neq s_j$, $s_1=x_1$, $s_L=x_N$. We define the \textit{piecewise linear cdf} $PL_S(x)$ obtained by 'drawing' the line segments from $(s_i, F_X(s_i))$ to $(s_j, F_X(s_j))$. Also, we assume that $PL_S(x)=0$ if $x < s_1$ and $PL_S(x)=1$ if $x\geq s_L$.

It is important to note that using a piecewise linear cdf $PL_S(x)$ as data model, we make the assumption that the subset $X(s_i,s_{i+1})$ of data points  in each interval $[s_i,s_{i+1}]$ is uniformly distributed. Thus $PL_S(x)$ is actually the cdf of a uniform mixture model (UMM). 

In UU-test, we aim to approximate the ecdf $F_X(x)$ using a $PL_S(x)$ that is unimodal. In order for the $PL_S(x)$ to be a good approximation of the ecdf, it should be \textit{sufficient} in the sense defined as follows:

Let a subset $S=\{s_1,...,s_L\} \subseteq X$ with $s_i \neq s_j$, $s_1=x_1$, $s_L=x_N$. Subset $S$ will be called \textit{sufficient} if the cdf $PL_S(x)$ is a good statistical model of $X$. Since $PL_S(x)$ models the data in each interval using the uniform distribution, in order for $PL_S(x)$ to be a good statistical model of $X$, for each $i$ the subset $X(s_i,s_{i+1})$ should follow the uniform distribution as decided by a uniformity test. Thus in the case where $PL_S(x)$ is sufficient, the corresponding uniform mixture model fits well to the data. 

If $PL_S(x)$ is both unimodal and sufficient then we consider that the dataset $X$ is unimodal and $PL_S(x)$ provides a good statistical model of $X$. Thus, \textit{the UU-test method searches for a subset $S$ of $X$, such that the cdf $PL_S(x)$ is unimodal and sufficient}.

In order to address the unimodality issue of $PL_S(x)$ we confine our search to the gcm and lcm points of the ecdf, exploiting the idea used in the dip-test method \cite{hartigan1985dip} for computing the dip statistic. 

\begin{figure}[H]
\centering   
    \begin{subfigure}[b]{0.40\linewidth}
    \includegraphics[width=\linewidth]{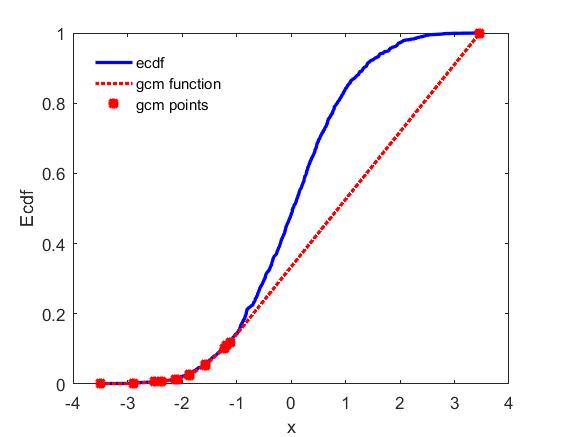}
    \caption{}
    \label{figure:fig2a}
  \end{subfigure}
 \begin{subfigure}[b]{0.40\linewidth}
    \includegraphics[width=\linewidth]{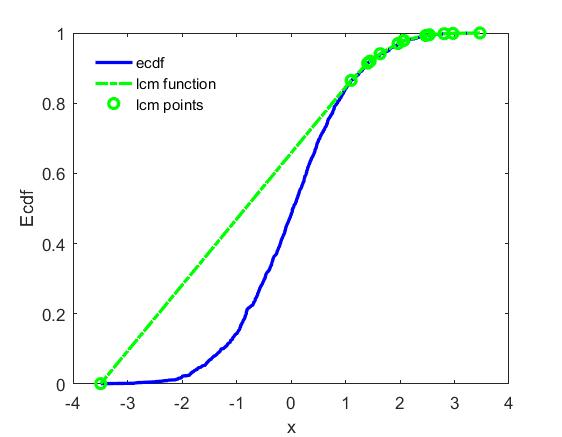}
    \caption{}
    \label{figure:fig2b}
  \end{subfigure}
  \caption{(a) Gcm function and gcm points of an ecdf. (b) Lcm function and lcm points of an ecdf.}
    \label{figure:fig2}
  \end{figure}

More specifically, we define the greatest convex minorant ($gcm$) of a function $F$ in $(-\infty,a]$ as $sup G(x)$ for $x\leq a$, where the $sup$ is taken over all functions $G$ that are convex in $(-\infty,a]$ and nowhere greater than $F$. Based on the above definition, we denote as $G_X(x)$ the gcm of ecdf $F_X(x)$. Note that, \textit{since $F_X(x)$ is piecewise constant, $G_X(x)$ is piecewise linear}. Let $G=\{g_1,...,g_{P_G}\} \subset X$ the set of gcm points (where $g_1=x_1, g_{P_G}=x_N$). The graph of $G_X(x)$ is defined by drawing line segments from ($g_i, F_X(g_i)$) to ($g_j, F_X(g_j)$). Based on the PL definition, we can write: $G_X(x)=PL_G(x)$. It should be mentioned that the gcm function of $F_X(x)$ corresponds to the monotonically increasing part of a pdf plot. Fig.~\ref{figure:fig2a} presents an ecdf, along with the gcm function $G_X(x)$ and the set of gcm points $G$.

Similarly, the least concave majorant ($lcm$) of a function $F$ in $[a,\infty)$ is defined as $inf L(x)$ for $x\geq a$,  where the $inf$ is taken over all functions $L$ that are concave in $[a,\infty)$ and nowhere less than $F$.

We denote as $L_X(x)$ the lcm of ecdf $F_X(x)$. Since $F_X(x)$ is piecewise constant, \textit{$L_X(x)$ is piecewise linear}. Let $L=\{l_1,...,l_{P_L}\} \subset X$ the set of lcm points (where $l_1=x_1, l_{P_L}=x_N$). Its graph is defined by drawing line segments from ($l_i, F_X(l_i)$) to ($l_j, F_X(l_j)$) and we can write that $L_X(x)=PL_L(x)$.
The lcm function of $F_X(x)$ corresponds to the monotonically decreasing part of a pdf plot. 
Fig.~\ref{figure:fig2b} presents an ecdf along with the lcm function $L_X(x)$ and the corresponding set of lcm points $L$.

Given the sets of gcm ($G$) and lcm points ($L$) of $F_X(x)$, we define as $GL$ the ordered set of points obtained from the union of $G$ and $L$:
$GL=\{v_1, ..., v_M\}$, where $v_1=x_1$, $v_M=x_N$, $v_i < v_j$ if $i < j$ and either $v_i \in G$ or $v_i \in L$. Note that $v_1=x_1$ and $v_M=x_N$ belong to both $G$ and $L$.
We also define as $maxG=max(v_i | v_i \in G-\{x_N\})$ and $minL=min(v_i | v_i \in L-\{x_1\})$, the maximum value of $G$ and the minimum value of $L$ respectively, excluding the maximum and minimum elements of $X$.

\begin{figure}[H]
	\centering
	\begin{subfigure}[b]{0.40\linewidth}
		\includegraphics[width=\linewidth]{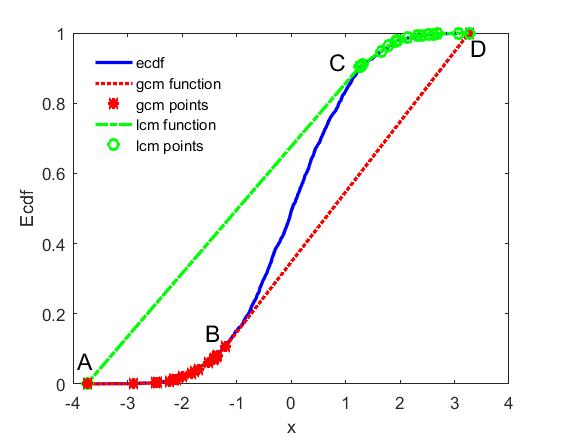}
	\end{subfigure}
	\caption{Gcm/Lcm function and gcm/lcm points of a unimodal ecdf. AB, BC and CD correspond to the convex, intermediate and concave part, respectively.}
	\label{figure:fig3}
\end{figure}

Let $S$ be a subset of $GL$ that i) includes $v_1$ and $v_M$ and ii) has the property that $maxG < minL$. Based on the definition of unimodality for cdf, it is straightforward to observe (see Fig.~\ref{figure:fig3}) that $PL_S(x)$ \textit{is unimodal} and we will call the set $S$ with the above two properties as \textit{consistent}. It should be stressed that this definition includes the cases where either the gcm or the lcm part is missing. 

A remarkable implication of consistency is that, \textit{since $PL_S(x)$ is unimodal, the set $S$ can be decomposed into three subsets} namely: 
\begin{itemize}
\item $S_G$ with the elements of $S$ less than or equal to $maxG$ (convex part, $PL_{S_G}(x)$ is convex) 
\item $P_I=\{maxG,minL\}$ (intermediate linear part, $PL_{P_I}(x)$ is linear)
\item $S_L$ with the elements of $S$ greater than or equal to $minL$ (concave part, $PL_{S_L}(x)$ is concave). 
\end{itemize}

Fig.~\ref{figure:fig3} presents a \textit{unimodal} ecdf and the gcm/lcm (GL) points. The three sets $S_G$, $P_I$ and $S_L$ correspond to segments AB, BC and CD, respectively.

Table~\ref{table:table1} summarizes the notations and definitions introduced in this section.
\vspace{2.3cm}
\begin{longtable}{|c|c|}\hline 
	Notation&Explanation \\\hline 
	$X=\{x_1,\ldots, x_N\}$ & A set of distinct N points in $\mathbb{R}.$ \\\hline
	$X(a,b)$  &The set $\{a \leq x_i \leq b, x_i\in X\}$ for an interval [a,b].  \\\hline
	$F_X(x)$ & Ecdf of $X$.  \\\hline
	$S$ & Subset of $X$. \\\hline
	$PL_S(x)$ & Piecewise linear cdf of $F_X(x)$.     \\\hline
	$G=\{g_1,...,g_{P_G}\}$   & The set  of gcm points of $F_X(x)$,  where $g_1=x_1, g_{P_G}=x_N$.        \\\hline
	$G_X(x)$   & The gcm function of $F_X(x)$.             \\\hline
	$L=\{l_1,...,l_{P_L}\}$   & The set of lcm points of $F_X(x)$, where $l_1=x_1, l_{P_L}=x_N$.         \\\hline
	$L_X(x)$   & The lcm function of $F_X(x)$.          \\\hline
	$GL$   & The set of ordered points of $G \cup L$.    \\\hline
	$maxG$   & $maxG=max(v_i | v_i \in G-\{x_N\})$.    \\\hline
	$minL$   & $minL=min(v_i | v_i \in L-\{x_1\})$.    \\\hline
	\textit{sufficient($S$)} & True, if $X(s_i,s_{i+1})$ is uniform for each $i$. \\\hline
	\textit{consistent($S$)} & True, if $S \subseteq GL$ and $S$ includes $x_1$ and $x_N$ and $maxG<minL$.    \\\hline
	\caption{Summary of notation.}
	\label{table:table1}
\end{longtable}

\section{UU-test description} \label{UU}

As mentioned in the previous section, UU-test aims at finding a subset $S$ of dataset $X$, such that the corresponding cdf $PL_S(x)$ is unimodal and sufficient. The latter means that the data in each interval $[s_i, s_{i+1}]$ are well-fitted by the uniform distribution. It should be noted that exhaustive search could have been used to determine an appropriate subset $S$, but it is computationally prohibitive. Alternatively, search techniques based on generate-and-test could also have been used. 

In the UU-test method, search is restricted to subsets $S$ of $GL=\{v_1,\ldots,v_M\}$, instead of examining the whole dataset $X$. We make the search even more focused, by looking for subsets of $GL$ that are consistent, since consistency implies unimodality. Thus, \textit{we search for a subset $S$ of $GL$ that is consistent and sufficient}. If such a set $S$ is found, then $PL_S(x)$ defines a unimodal distribution that sufficiently models the dataset $X$. In this case UU-test decides unimodality and outputs the corresponding statistical model. 

Given a 1-d dataset $X=\{x_1, \ldots, x_N\}$, function $S=UUtest(X)$ (Algorithm \ref{algo:algo1}) takes $X$ as input and outputs a non-empty set $S$ (that is consistent and sufficient) in the case of unimodality and the empty set $S=\emptyset$ in the case of multimodality. It first computes the ecdf of the dataset (set $E$) and then calls function $UU$ (Algorithm \ref{algo:algo2}) where most of the work takes place.
   
\begin{algorithm}[H]
	\caption{$S=UUtest(X)$}
	\label{algo:algo1}
	\begin{algorithmic}
		\State{$E=(x_i,F_X(x_i)) \leftarrow ecdf(X)$}
		\State{$S_G \leftarrow \emptyset$, \space $S_L \leftarrow \emptyset$, \space success $\leftarrow$ true, \space $P_I \leftarrow \{x_{min},x_{max}\}$}
		\State{$(S_G',P_I',S_L',success) \leftarrow UU(S_G,P_I,S_L)$}
		\State{\Return{$S \leftarrow S_G' \cup S_L'$}}
	\end{algorithmic}
\end{algorithm}

\begin{algorithm}[H]
\caption{$(S_G',P_I',S_L',success)=UU(S_G,P_I,S_L)$}
\label{algo:algo2}
\begin{algorithmic}
\If{$check{\_}uniformity(X(P_I))=true$} 
\State{\Return{$(S_G,P_I,S_L,true)$}}
\EndIf
\State{$E_I=\{(x_i,y_i) \in E/ x_i \in X(P_I)\}$}
\State{$GL \leftarrow$ \textbf{compute} gcm \& lcm points of $E_I$}
\State{\textbf{determine} set $C$ of consistent subsets of $GL$}
\ForEach{consistent subset $GL_C \in C$} 
\State{($P_G',P_I',P_L') \leftarrow $decompose($GL_C$)}
\State{$(S_G',success) \leftarrow $sufficient$(P_G')$}
\If{success=false} \State{\textbf{continue}}
\EndIf
\State{$(S_L',success) \leftarrow $sufficient$(P_L')$}
\If{success=false} \State{\textbf{continue}}
\EndIf
\State{$S_G' \leftarrow S_G'\cup S_G$}
\State{$S_L' \leftarrow S_L'\cup S_L$}
\State{$(S_G'',P_I'',S_L'',success) \leftarrow UU(S_G',P_I',S_L')$}
\If{success=true} \State{\Return{$(S_G'',P_I'',S_L'',true)$}}
\EndIf
\EndFor
\State{\Return{($\emptyset,\emptyset,\emptyset,$false)}}
\end{algorithmic}
\end{algorithm}

$UU$ function takes three sets as input, namely $S_G$ (convex part), $P_I$ (intermediate part) and $S_L$ (concave part) and, if successful, it returns (possibly) updated versions of the three sets, otherwise it returns empty sets. Initially $S_G$ and $S_L$ are empty, while $P_I=\{x_1, x_N\}$, i.e., $X(P_I)=X(x_1,x_N)=X$. $UU$ function operates on the data in the intermediate part $X(P_I)$. At first it checks for early success, this means that we test the uniformity of $X(P_I)$. If this happens, the function terminates successfully. 

\subsection{Consistent Subsets}
If $X(P_I)$ is not uniform, we compute the corresponding set $GL$ (union of gcm and lcm points) of $X(P_I)$ and determine the set $C$ containing the consistent subsets $GL_C$ of $GL$. Two cases are considered: 
\begin{itemize}
	\item {either $C=\{GL\}$, i.e. $GL$ is itself consistent, $maxG<minL$}
	\item{or $C=\{GL_1, GL_2\}$}
\end{itemize}
In the latter case, the first consistent subset  ($GL_1$) is obtained by removing all gcm points that lie after the first lcm point. Similarly, the second consistent subset ($GL_2$) is obtained by removing all lcm points that lie before the last gcm point.

Next we examine each set $GL_C\in C$. Since $GL_C$ is consistent, it is decomposed into three sets corresponding to the convex ($P'_G$), intermediate ($P'_I$) and concave $(P'_L$) part. Then we try to determine a sufficient subset $S'_G$ of $P'_G$ as well as a sufficient subset $S'_L$ of $P'_L$. In the case of failure, the second consistent subset $GL_C$ is examined (if it exists). In the case of success (i.e. both sufficient sets $S'_G$ and $S'_L$ have been found), the sets $S'_G$ and $S'_L$ are updated, and the $UU$ function is called recursively in order to examine the intermediate part $P'_I$. The recursion ends either if $P'_I$ cannot be decomposed into a sufficient gcm part ($S''_G$) and a sufficient lcm part ($S''_L$) (unsuccessful termination) or when $X(P'_I)$ is found uniform (successful termination). If $UU$ is successfully applied on $X(P'_I)$ providing the sets $S''_G,P''_I ,S''_L$, then $S'' = S''_G \cup P''_I \cup S''_L$ is the final solution for $X$. If the $UU$ function fails on $X(P'_I)$, then the calling function $UU(X(P_I))$ also fails for the specific consistent subset $GL_C$. In this case the second $GL_C$ subset (if it exists) should be examined.

Fig.~\ref{figure:fig4} concerns a multimodal dataset. The histogram and the ecdf are presented in Fig.~\ref{figure:fig4a} and Fig.~\ref{figure:fig4b} respectively. In Fig.~\ref{figure:fig4b} the GL points are also presented. It can be observed that there exist lcm points (e.g. A) that lie before a gcm point (B). Therefore GL is inconsistent. In Fig.~\ref{figure:fig4c} we consider the consistent subset of GL that is obtained by omitting the lcm points (e.g. A) that lie between gcm points (B) and (C). Another consistent subset of GL can be obtained by omitting the gcm point (B) that lies between lcm points A and D. This case is shown in Fig.~\ref{figure:fig4d}. In $UU$ function, both consistent subsets are checked for sufficiency and they fail in this test. Thus the dataset is characterized as multimodal.
 
\begin{figure}[H]
\centering 
    \begin{subfigure}[b]{0.4\linewidth}
    \includegraphics[width=\linewidth]{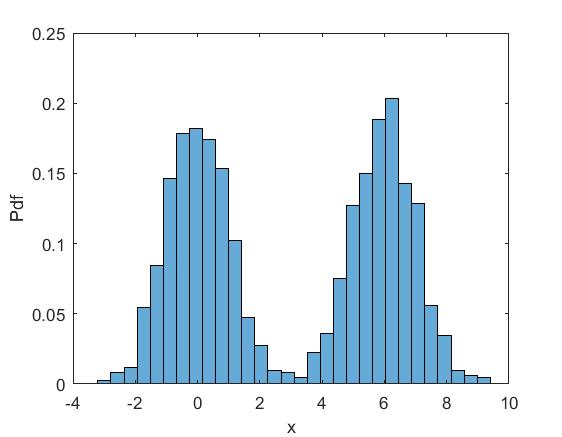}
    \caption{}
    \label{figure:fig4a}
  \end{subfigure}
 \begin{subfigure}[b]{0.4\linewidth}
    \includegraphics[width=\linewidth]{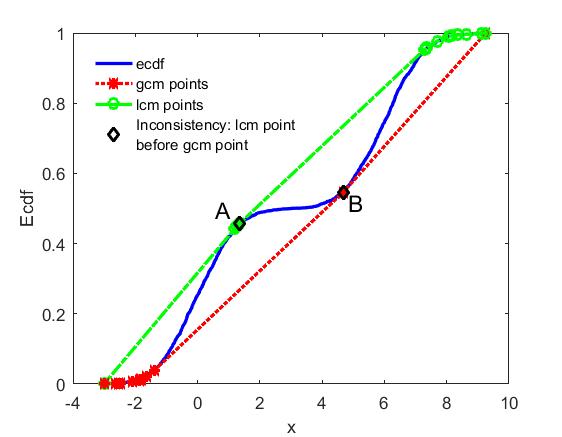}
    \caption{}
    \label{figure:fig4b}
  \end{subfigure}  
  \end{figure}
\vspace{-0.5cm}
  \begin{figure}[H]\ContinuedFloat
\centering 
    \begin{subfigure}[b]{0.4\linewidth}
    \includegraphics[width=\linewidth]{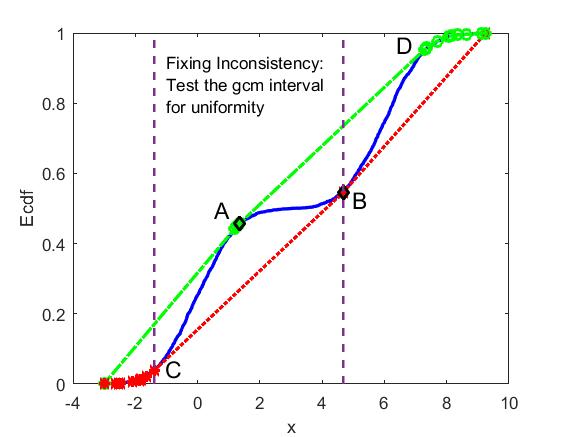}
    \caption{}
    \label{figure:fig4c}
    \end{subfigure}
    \begin{subfigure}[b]{0.4\linewidth}
    \includegraphics[width=\linewidth]{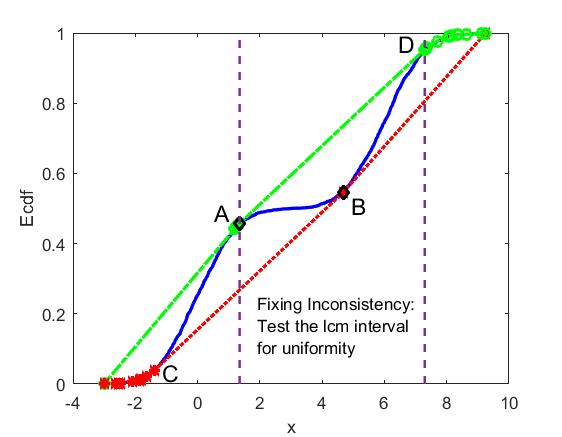}
    \caption{}
    \label{figure:fig4d}
  \end{subfigure}
\caption{Example of multimodal dataset with consistent GL subsets that are not sufficient.}
\label{figure:fig4}
\end{figure}

Fig.~\ref{figure:fig5} concerns a unimodal dataset. The histogram and the ecdf are presented in Fig.~\ref{figure:fig5a} and Fig.~\ref{figure:fig5b} respectively. In Fig.~\ref{figure:fig5b} the GL points are also presented. It can be observed that there exists a lcm point (A) that lies before gcm points (e.g. B). Therefore GL is inconsistent. In Fig.~\ref{figure:fig5c} we consider the consistent subset of GL that is obtained by omitting the lcm point (A) that lies between gcm points (B) and (C). Another consistent subset of GL can be obtained by omitting the gcm points (e.g. B) that lie between lcm points A and D. This case is shown in Fig.~\ref{figure:fig5d}. In contrast to the case of Fig.~\ref{figure:fig4}, both consistent subsets are sufficient. In $UU$ function, the first consistent subset is checked for sufficiency and succeeds in this test. Thus the dataset is characterized as unimodal.

 \begin{figure}[H]
\centering 
    \begin{subfigure}[b]{0.4\linewidth}
    \includegraphics[width=\linewidth]{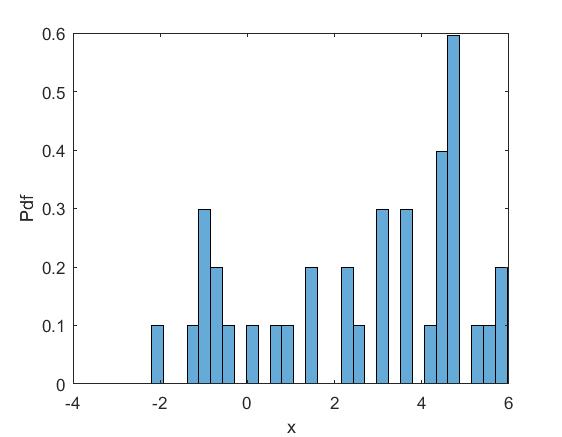}
    \caption{}
    \label{figure:fig5a}
  \end{subfigure}
 \begin{subfigure}[b]{0.4\linewidth}
    \includegraphics[width=\linewidth]{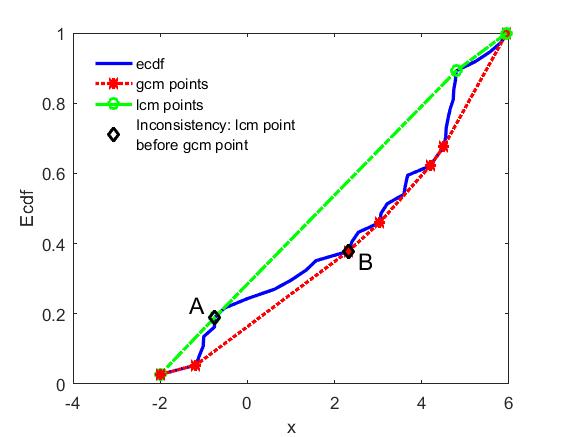}
    \caption{}
    \label{figure:fig5b}
  \end{subfigure}
    \end{figure}
\vspace{-0.5cm}
  \begin{figure}[H]\ContinuedFloat
\centering 
    \begin{subfigure}[b]{0.4\linewidth}
    \includegraphics[width=\linewidth]{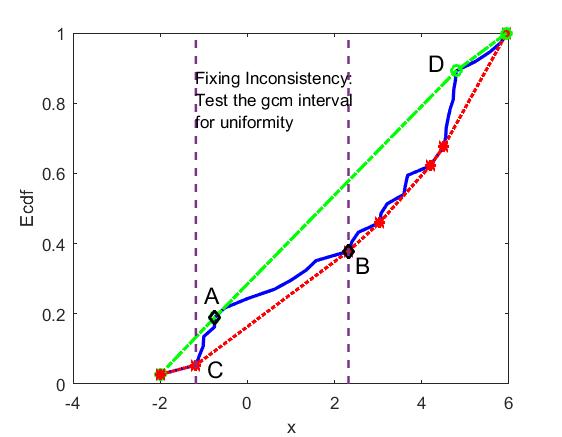}
    \caption{}
    \label{figure:fig5c}
    \end{subfigure}
    \begin{subfigure}[b]{0.4\linewidth}
    \includegraphics[width=\linewidth]{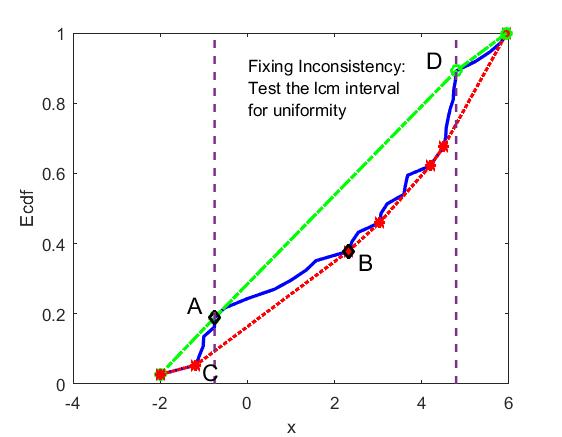}
    \caption{}
    \label{figure:fig5d}
  \end{subfigure}
\caption{Example of unimodal dataset with consistent GL subsets that are sufficient.}
\label{figure:fig5}
  \end{figure}

\subsection{Sufficient subsets}

Let $GL_C=\{s_1, \ldots,s_K\}$ be a consistent subset of $GL$. Note that $s_1=x_1$ and $s_K=x_N$. Since $GL_C$ is consistent,  $PL_{GL_C}(x)$ is unimodal. In the general case, $GL_C$ contains both gcm and lcm points. Thus, there exists a single index $c$ such that all elements $s_i$ for $i=1,\ldots,c$ are gcm points and all elements $s_j$ (for $j=c+1,\ldots,K$) are lcm points.  Thus we can write that: $GL_C=P_G \cup P_I \cup P_L$, where $P_G=\{s_1,\ldots,s_c\}$ (gcm elements), $P_I= \{s_c, s_{c+1}\}$ and $P_L=\{s_{c+1},\ldots,s_K\}$ (lcm elements). 

Moreover, \textit{every subset of $GL_C$ that includes the points $s_1, s_c, s_{c+1},s_K$ is also consistent} (i.e. cdf $PL_{S}$ is unimodal). Therefore it can be decomposed into three parts. The convex part which is a subset of $P_G$ having $s_1$ and $s_c$ as the first and last element. The concave part which is a subset of $P_L$ having  $s_{c+1}$ and $s_K$ as the first and last element. The intermediate part is always the two-element set $P_I=\{s_c, s_{c+1}\}$. Thus our objective is to find a subset $S$ of $GL_C$ that is also sufficient. 

In order to determine a sufficient subset $S'_G$ of $P_G=\{s_1,\ldots,s_c\}$ (convex part) we work as follows: We first test whether the subset $X(s_1, s_c)$ succeeds in the uniformity test. If this is the case, we have successfully determined a sufficient set $S_G'=\{s_1, s_c\}$. However, if it fails, we continually test the successive subsets $X(s_i,s_{i+1})$ $(i=1, \ldots, c-1)$ using the uniformity test. If the test succeeds, the points $s_i,s_{i+1}$ are saved in $S'_G$. If all subsets $X(s_i, s_{i+1})$ are uniform, then $P_G$ is sufficient ($S'_G=P_G$).  However, it is possible for some subset to fail in the uniformity test. Let $X(s_i,s_{i+1})$ the first non-uniform data subset that is encountered, thus currently $S'_G=\{s_1,\ldots, s_i\}$. We make two attempts to fix this problem.

The first attempt is the Forward search method that searches for uniform supersets of $X(s_i,s_{i+1})$ by \textit{moving the right interval endpoint}, i.e. $X(s_i,s_j)$, $j>i+1$. This method works in increasing order of $s_i$ and successively tests whether the sets $X(s_i,s_{i+2})$, $X(s_i,s_{i+3})$, $X(s_i,s_{i+4})$ etc. are uniform. If a set  $X(s_i,s_j)$, $j>i+1$ is found  uniform, the element $s_j$ is added in $S_G'$ and we continue by testing the next subset $X(s_j,s_{j+1})$ for uniformity. 

If the Forward search fails, the Backward search method is called that searches for uniform supersets of $X(s_i,s_{i+1})$ by \textit{moving the left interval endpoint}, i.e. $X(s_m,s_{i+1})$, $m<i$. This method searches backwards and tests successively if the sets $X(s_{i-1},s_{i+1})$, $X(s_{i-2},s_{i+1})$, $X(s_{i-3},s_{i+1})$ etc. are uniform. If such a set is found, the non-uniformity problem is fixed. More specifically, if a set  $X(s_m,s_{i+1})$, $m<i$ is found  uniform, the elements $s_{m+1}, \ldots, s_i$ are removed from $S_G'$ and we continue by testing if the next subset $X(s_{i+1},s_{i+2})$ succeeds in the uniformity test.

In order to determine a sufficient subset $S’_L$ (concave part)  we work in a similar way with the $S_G'$ set. Algorithm~\ref{algo:algo3} describes the overall method of determining a sufficient subset of a convex or concave set. We denote $e_n=$ next$(e,P)$ the next element of $e$ in set $P$ and $e_p=$ prev$(e,P)$ the previous element of $e$ in set $P$. Algorithm~\ref{algo:algo4} describes the Forward search method, while Algorithm~\ref{algo:algo5} describes the steps of the Backward search method.

Fig.~\ref{figure:fig6} presents an example of a multimodal dataset which exhibits non-uniformity in the interval between two successive lcm points. Fig.~\ref{figure:fig6a} presents the histogram of the dataset and Fig.~\ref{figure:fig6b} the ecdf of the dataset along with the GL points. It can be observed that the part of the ecdf between lcm points A and B is not linear, i.e. the subset is not uniform. In Fig.~\ref{figure:fig6c} and Fig.~\ref{figure:fig6d} we zoom into the concave (lcm) part of the ecdf where the nonlinearity (i.e. non-uniformity) of the ecdf is made more clear. In such a case we attempt to fix this issue by using the Forward and Backward search algorithms, however in this example both attempts fail.

Fig.~\ref{figure:fig7} concerns an example of a unimodal dataset that includes a data subset in the  concave part that is not uniform. However, in contrast to the case of Fig.~\ref{figure:fig6}, the Forward search algorithm manages to fix this problem. Fig.~\ref{figure:fig7a} presents the histogram of the dataset and Fig.~\ref{figure:fig7b} the ecdf of the dataset along with the GL points. It can be observed that the ecdf segment between successive lcm points A and B is not linear. In Fig.~\ref{figure:fig7c} and Fig.~\ref{figure:fig7d} we zoom into the lcm part of the dataset. Fig.~\ref{figure:fig7c} presents the histogram and Fig.~\ref{figure:fig7d} the ecdf of this subset, where subset $X(A,B)$ between points A and B is characterized non-uniform. Using the Forward search algorithm, the superset $X(A,C)$ is found uniform, thus the non-uniformity issue is fixed.  

\begin{figure}[H]
\centering  
    \begin{subfigure}[b]{0.4\linewidth}
    \includegraphics[width=\linewidth]{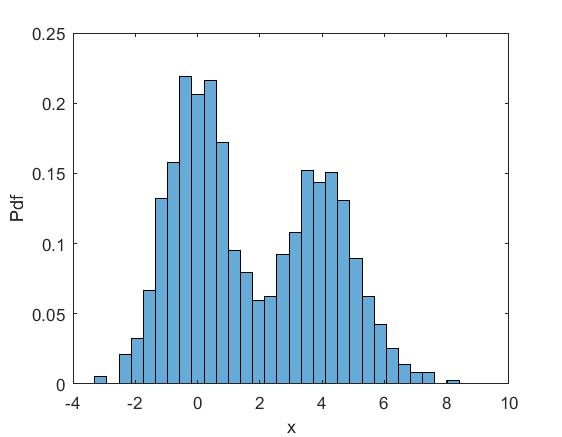}
    \caption{}
    \label{figure:fig6a}
  \end{subfigure}
 \begin{subfigure}[b]{0.4\linewidth}
    \includegraphics[width=\linewidth]{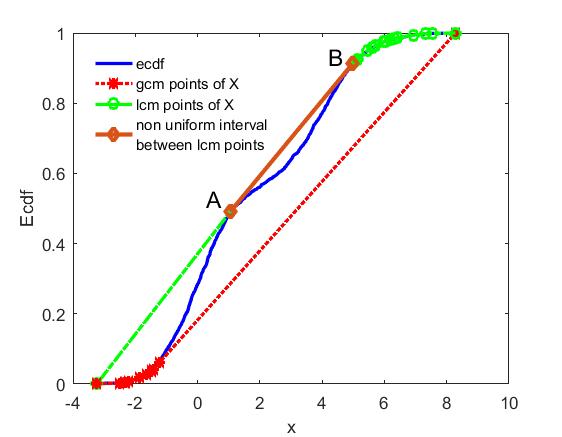}
    \caption{}
    \label{figure:fig6b}
  \end{subfigure}
    \end{figure}
\vspace{-0.5cm}
  \begin{figure}[H]\ContinuedFloat
\centering  
    \begin{subfigure}[b]{0.4\linewidth}
    \includegraphics[width=\linewidth]{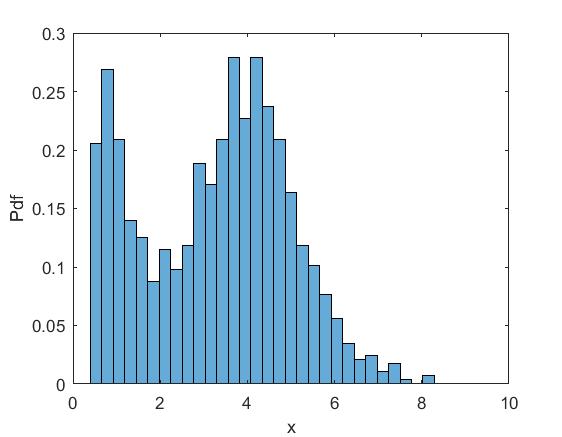}
    \caption{}
    \label{figure:fig6c}
    \end{subfigure}
    \begin{subfigure}[b]{0.4\linewidth}
    \includegraphics[width=\linewidth]{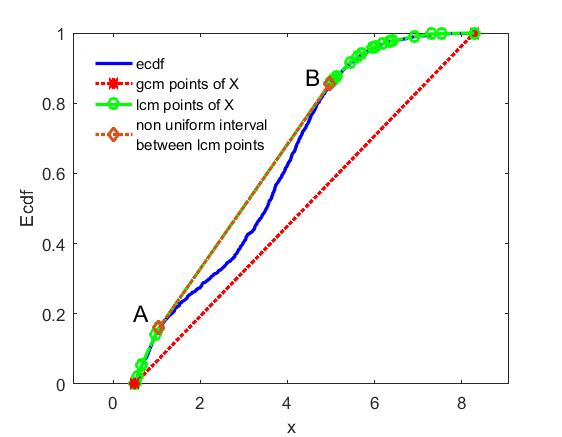}
    \caption{}
    \label{figure:fig6d}
  \end{subfigure}
\caption{Example of multimodal dataset where consistent subsets are not sufficient.}
\label{figure:fig6}
\end{figure} 

Two examples of the recursive application of $UU$ function are presented in Fig.~\ref{figure:fig8} and Fig.~\ref{figure:fig9}. Fig.~\ref{figure:fig8} concerns a multimodal dataset whose histogram is shown in Fig.~\ref{figure:fig8a}. In Fig.~\ref{figure:fig8b} the ecdf is presented along with the gcm and lcm points (GL points). It can be observed that the intermediate part $X(A,B)$ (between points A and B) of the ecdf is not linear (uniform). Fig.~\ref{figure:fig8c} and Fig.~\ref{figure:fig8d} focus on the intermediate part presenting the histogram of this subset and the ecdf respectively. In Fig.~\ref{figure:fig8d} it is clear that the intermediate part is not uniform. For this reason the $UU$ function is recursively applied on subset $X(A,B)$. Fig.~\ref{figure:fig8d} presents the $GL$ points of $X(A,B)$. It can be observed that there exist lcm points among gcm points, $UU$ function cannot fix this inconsistency, thus the whole dataset is characterized as multimodal. In Fig.~\ref{figure:fig8e} the initial ecdf is presented along with both the $GL$ points of the initial ecdf and the $GL$ points of the ecdf of the intermediate part. It is clear that there exist gcm points that lie among lcm points and this observation leads to decide multimodality.  

\begin{figure}[H]
\centering  
    \begin{subfigure}[b]{0.4\linewidth}
    \includegraphics[width=\linewidth]{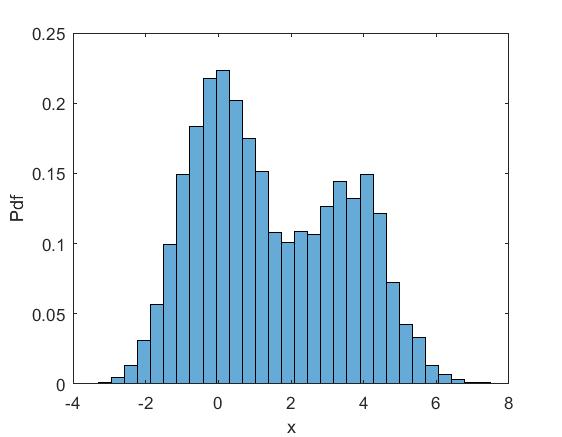}
    \caption{}
    \label{figure:fig7a}
  \end{subfigure}
 \begin{subfigure}[b]{0.4\linewidth}
    \includegraphics[width=\linewidth]{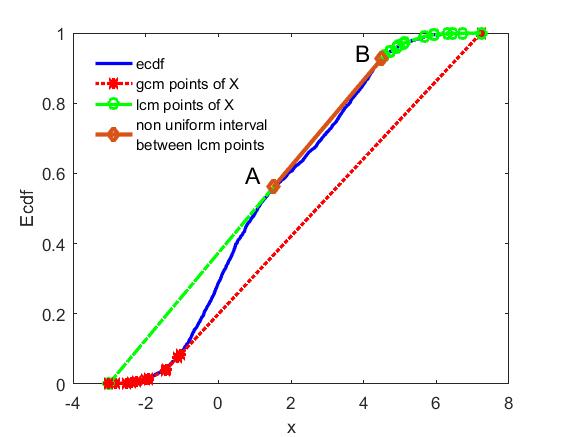}
    \caption{}
    \label{figure:fig7b}
  \end{subfigure}
  \end{figure}
\vspace{-0.5cm}
  \begin{figure}[H]\ContinuedFloat
\centering  
    \begin{subfigure}[b]{0.4\linewidth}
    \includegraphics[width=\linewidth]{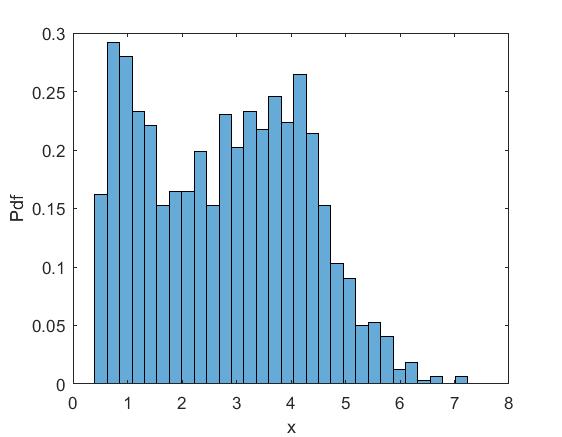}
    \caption{}
    \label{figure:fig7c}
    \end{subfigure}
    \begin{subfigure}[b]{0.4\linewidth}
    \includegraphics[width=\linewidth]{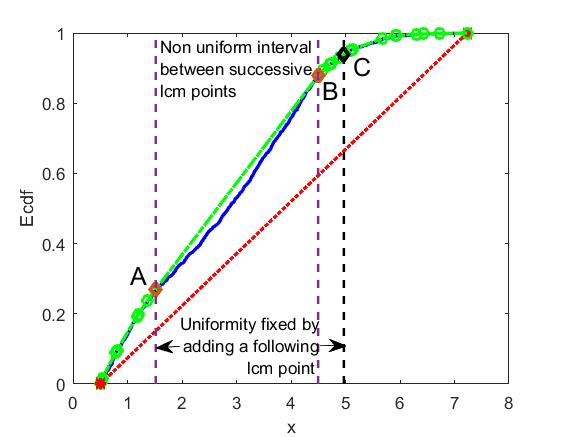}
    \caption{}
    \label{figure:fig7d}
  \end{subfigure}
\caption{Example of unimodal dataset. A non-uniform interval exists between lcm points A and B. Forward search method fixes the non-uniformity problem by considering the extended interval between A and C.}
\label{figure:fig7}
  \end{figure}

\hspace{-2em}
\begin{minipage}{0.51\textwidth} 
\begin{algorithm}[H] 
\caption{($P'$,success)=sufficient($P$)}
\label{algo:algo3}
\begin{algorithmic} 
\State{$e_1 \leftarrow$ min($P$)}
\State{$P' \leftarrow \{e_1\}$, success $\leftarrow$ true}
\While{$e_L \leftarrow$ max$(P') \neq max(P)$}
\If{$e_R \leftarrow$ next$(e_L,P)$ not exist}
\State{\Return{$(\emptyset,false)$}}
\EndIf
\If{check{\_}uniformity(X($e_L,e_R$))=true}
\State{$P' \leftarrow P' \cup \{e_R\} $}
\Else
\State{($P_F'$,success) $\leftarrow$ Forward{\_}search($P,e_L$)}
\If{success=true}
\State{$P' \leftarrow P' \cup P_F'$}
\Else
 \State{($P_B'$,success)$\leftarrow$Backward{\_}search($P',e_R$) }
\If{success=false}
\State{\Return{($\emptyset,false$)}}
\EndIf
\State{$P' \leftarrow P_B'$}
\EndIf
\EndIf
\EndWhile
\State{\Return{($P'$,success)}}
\end{algorithmic}
\end{algorithm}
\end{minipage} 
\hfill
\hspace{2em}
\begin{minipage}{0.49\textwidth}
\begin{algorithm}[H]
\caption{($P_F'$,success)=Forward{\_}search($P_F,e_L$)}
\label{algo:algo4}
\begin{algorithmic}
\State{$P_F\leftarrow P_F-\{$next$(e_L,P_F)\}$,\space$e_R\leftarrow$next$(e_L,P_F)$}
\While{$e_R$ exist}
\If{check{\_}uniformity(X($e_L,e_R$))=true}
\State{$P_F' \leftarrow \{e_R\}$, \space \Return{($P_F'$, true)}}
\EndIf
\State{$e_R \leftarrow$ next$(e_R,P_F)$}
\EndWhile
\State{\Return{($\emptyset$,false)}}
\end{algorithmic}
\end{algorithm}
\vspace{-2mm}
\begin{algorithm}[H]
\caption{($P_B'$,success)=Backward{\_}search($P_B,e_R$)}
\label{algo:algo5}
\begin{algorithmic}
\State{$P_B' \leftarrow P_B-\{$maximum element of $P_B$$\}$}
\State{$e_L \leftarrow$ max$(P_B')$}
\While{$e_L$ exist}
\If{check{\_}uniformity(X($e_L,e_R$))=true}
\State{$P_B' \leftarrow P_B' \cup \{e_R\}$, \space \Return{($P_B'$, true)}}
\EndIf
\State{$P_B' \leftarrow P_B' - \{e_L\}$, $e_L \leftarrow$ prev$(e_L,P_B')$}
\EndWhile
\State{\Return{($\emptyset$,false)}}
\end{algorithmic}
\end{algorithm}
\end{minipage}

\begin{figure}[H]
\centering   
    \begin{subfigure}[b]{0.4\linewidth} 
    \includegraphics[width=\linewidth]{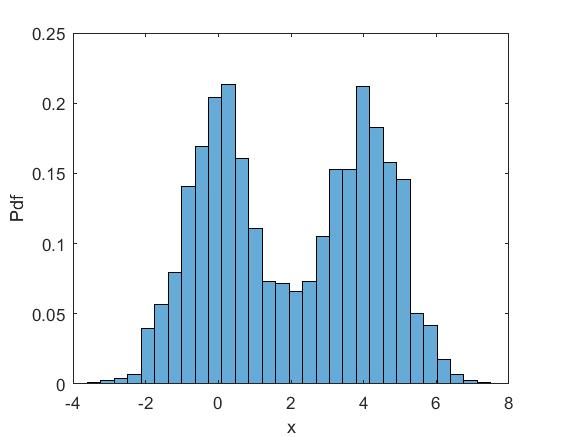}
    \caption{}
    \label{figure:fig8a}
  \end{subfigure}
 \begin{subfigure}[b]{0.4\linewidth}
    \includegraphics[width=\linewidth]{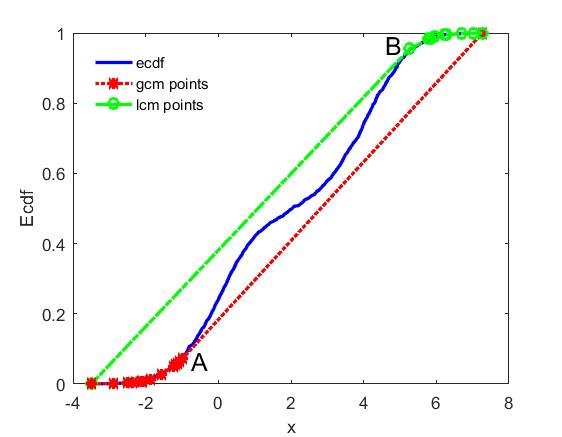}
    \caption{}
    \label{figure:fig8b}
  \end{subfigure}
    \end{figure}
\vspace{-0.5cm}
  \begin{figure}[H]\ContinuedFloat
\centering  
    \begin{subfigure}[b]{0.4\linewidth}
    \includegraphics[width=\linewidth]{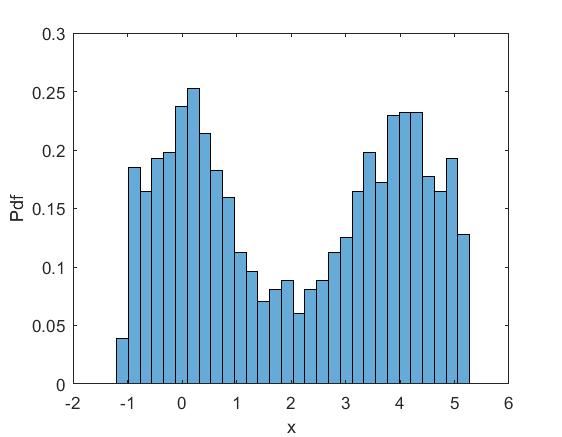}
    \caption{}
    \label{figure:fig8c}
    \end{subfigure}
    \begin{subfigure}[b]{0.4\linewidth}
    \includegraphics[width=\linewidth]{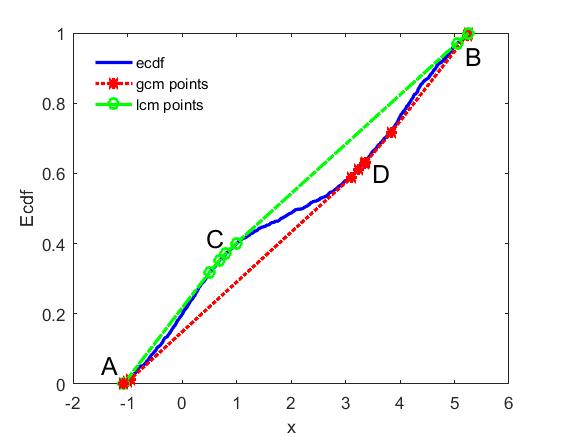}
    \caption{}
    \label{figure:fig8d}
  \end{subfigure}
  \end{figure}
\vspace{-0.5cm}
  \begin{figure}[H]\ContinuedFloat
   \centering
      \begin{subfigure}[b]{0.4\linewidth}
    \includegraphics[width=\linewidth]{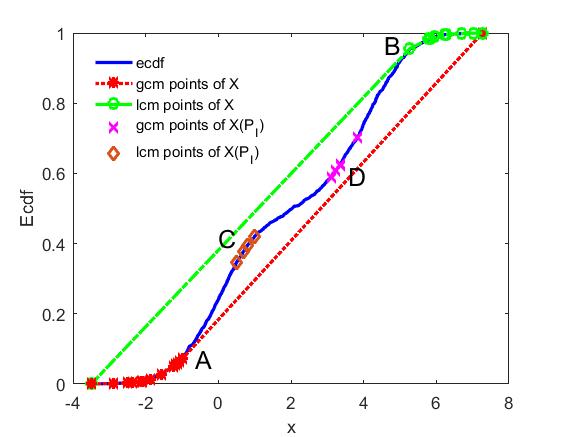}
    \caption{}
    \label{figure:fig8e}
  \end{subfigure}
\caption{Example of multimodal dataset where the UU function is recursively applied on the intermediate part.}
\label{figure:fig8}
  \end{figure}

Fig.~\ref{figure:fig9} concerns a unimodal dataset whose histogram is shown in Fig.~\ref{figure:fig9a}. In Fig.~\ref{figure:fig9b} the ecdf is presented along with the gcm and lcm points (GL points). It can be observed that the intermediate part of the ecdf is not linear (uniform). Fig.~\ref{figure:fig9c} and Fig.~\ref{figure:fig9d} focus on the intermediate part $X(A,B)$ presenting the histogram of this subset and the ecdf respectively. In Fig.~\ref{figure:fig9d} it is clear that the intermediate part is not uniform and $UU$ function is recursively applied on this subset. As shown in Fig.~\ref{figure:fig9d}, the intermediate part is unimodal and the whole dataset is characterized as unimodal. In Fig.~\ref{figure:fig9e} the initial ecdf is presented along with both the GL points of the initial ecdf and the GL points of the ecdf of the intermediate part. It can be observed that all gcm points precede the lcm points and this is an indication of unimodality, provided that the sufficiency criterion is also met. 

 \begin{figure}[H]
\centering  
    \begin{subfigure}[b]{0.4\linewidth}
    \includegraphics[width=\linewidth]{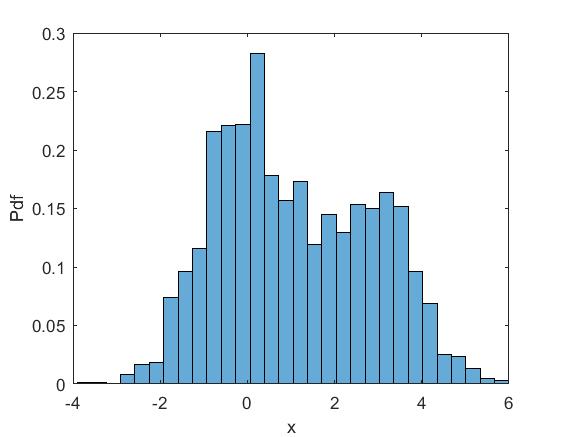}
    \caption{}
    \label{figure:fig9a}
  \end{subfigure}
 \begin{subfigure}[b]{0.4\linewidth}
    \includegraphics[width=\linewidth]{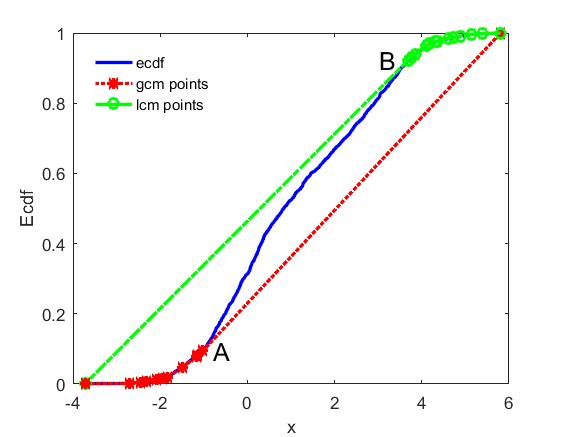}
    \caption{}
    \label{figure:fig9b}
  \end{subfigure}
    \end{figure}
\vspace{-0.5cm}
  \begin{figure}[H]\ContinuedFloat
\centering  
    \begin{subfigure}[b]{0.4\linewidth}
    \includegraphics[width=\linewidth]{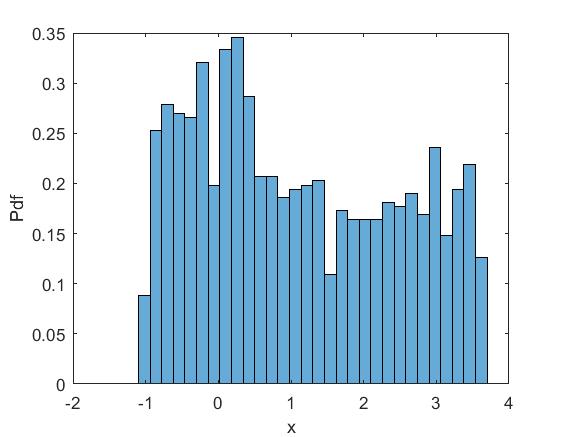}
    \caption{}
    \label{figure:fig9c}
    \end{subfigure}
    \begin{subfigure}[b]{0.4\linewidth}
    \includegraphics[width=\linewidth]{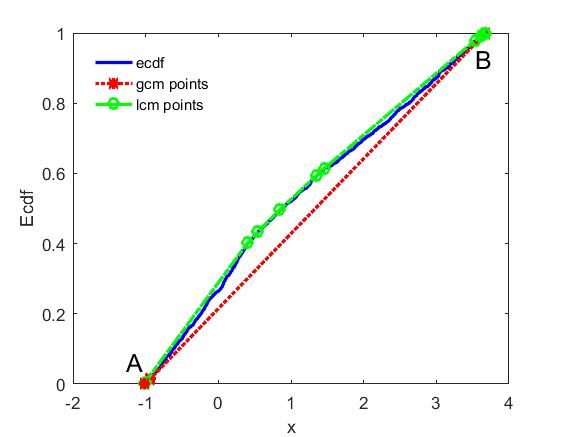}
    \caption{}
    \label{figure:fig9d}
  \end{subfigure}
    \end{figure}
\vspace{-0.5cm}
  \begin{figure}[H]\ContinuedFloat
   \centering
      \begin{subfigure}[b]{0.4\linewidth}
    \includegraphics[width=\linewidth]{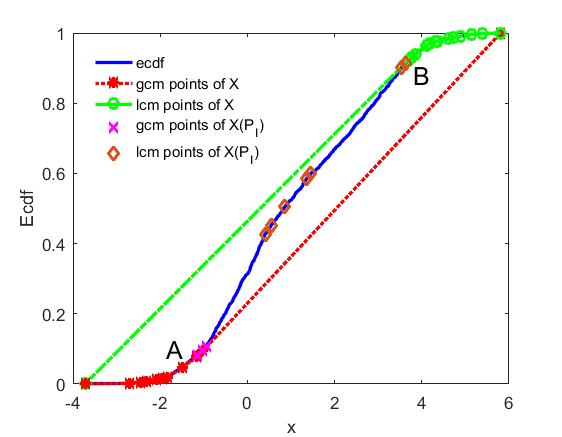}
    \caption{}
    \label{figure:fig9e}
  \end{subfigure}
\caption{Example of unimodal dataset where the UU function is recursively applied on the intermediate part.}
\label{figure:fig9}
  \end{figure}

\subsection{Uniformity Test}
A very common operation in the UU-test method, is to decide whether a subset is sufficiently modeled by the uniform distribution. For this reason a uniformity test is needed. In our implementation we use the Kolmogorov-Smirnov test (KS test) as a uniformity test. KS test first computes the KS statistic, which is the distance between the ecdf of the dataset and the cdf of the uniform distribution. Next a $p$-value is determined and compared with a user-defined significance level $\alpha$ (we use $\alpha=0.01$ in our experiments). Therefore, if $p$-value $\leq a$, the KS test will reject uniformity. 

There are two interesting features of KS test. First, the distribution of the KS test statistic itself does not depend on the underlying cumulative distribution function being tested and second, it is an exact test. Moreover, it is straightforward to determine the corresponding $p$-value, while the dip-test employs bootstrapping to compute the $p$-value.   

However, KS test exhibits a peculiarity that may affect the result. It tends to be more sensitive near the center of the distribution than at the tails. In several experiments with large unimodal datasets, the KS test fails to early accept the uniformity of the intermediate part requiring the additional iterations. However, the final unimodality decision is not affected.

\subsection{Computational complexity}
The computational complexity of UU-test mainly depends on cost of computing the gcm/lcm points of the ecdf which is O(n) using the isotonic regression method for gcm points and antitonic regression for lcm points \cite{robertson1988order}.
Gcm/lcm points can also be determined in O(nlogn) from the convex hull of the ecdf plot. If the data are not sorted an additional O(nlogn) complexity should be considered. It should be stressed that UU-test relies on KS-test, thus it does not require extra computations on bootstrap samples to obtain the p-value. 

\section{Modeling Unimodal Data} \label{Modeling}

\begin{figure}[H]
\centering 
	\begin{subfigure}[b]{0.4\linewidth}
		\includegraphics[width=\linewidth]{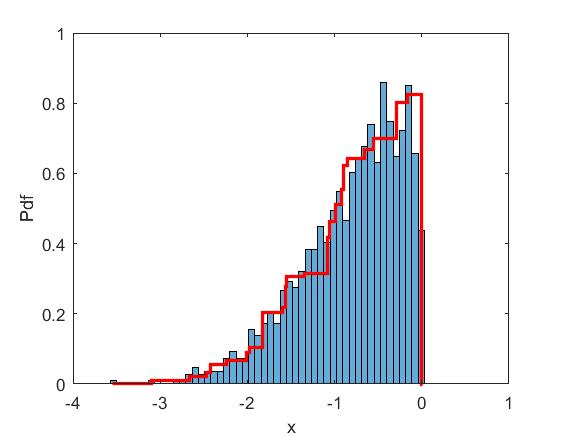}
		\caption{}
		\label{figure:fig10a}
	\end{subfigure}
	\begin{subfigure}[b]{0.4\linewidth}
		\includegraphics[width=\linewidth]{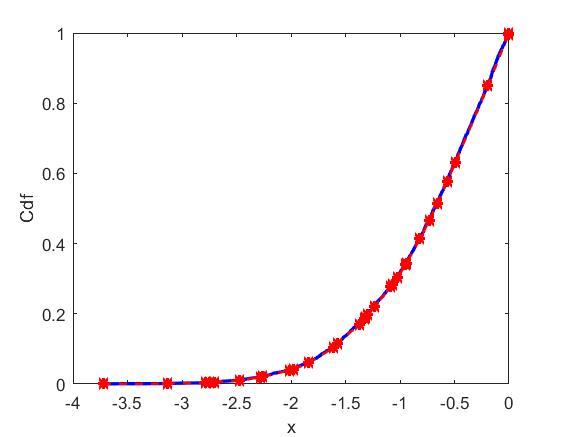}
		\caption{}
		\label{figure:fig10b}
	\end{subfigure}
	\caption{Dataset sampled from a truncated ($x<0$) Gaussian distribution. (a) Histogram and UMM pdf (solid line). (b) Points of $S$, ecdf (solid line) and UMM cdf (dashed line).}
	\label{figure:fig10}
\end{figure}

\begin{figure}[H]
\centering	
	\begin{subfigure}[b]{0.4\linewidth}
		\includegraphics[width=\linewidth]{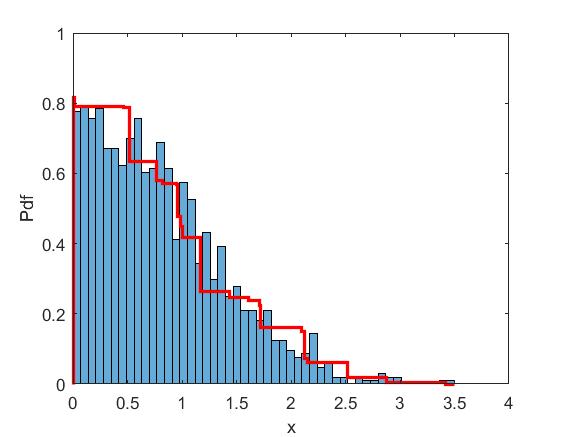}
		\caption{}
		\label{figure:fig11a}
	\end{subfigure}
	\begin{subfigure}[b]{0.4\linewidth}
		\includegraphics[width=\linewidth]{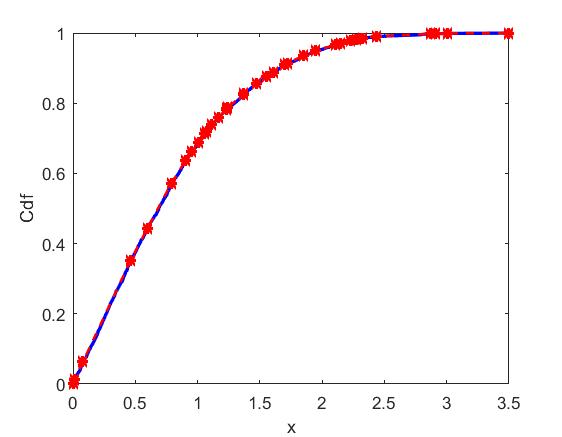}
		\caption{}
		\label{figure:fig11b}
	\end{subfigure}
	\caption{Dataset sampled from a truncated ($x>0$) Gaussian distribution. (a) Histogram and UMM pdf (solid line). (b) Points of $S$, ecdf (solid line) and UMM cdf (dashed line).}
	\label{figure:fig11}
\end{figure}

\begin{figure}[H]
\centering	 
	\begin{subfigure}[b]{0.4\linewidth}
		\includegraphics[width=\linewidth]{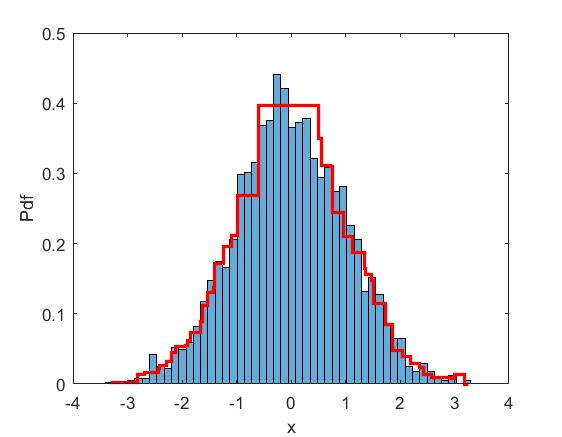}
		\caption{}
		\label{figure:fig12a}
	\end{subfigure}
	\begin{subfigure}[b]{0.4\linewidth}
		\includegraphics[width=\linewidth]{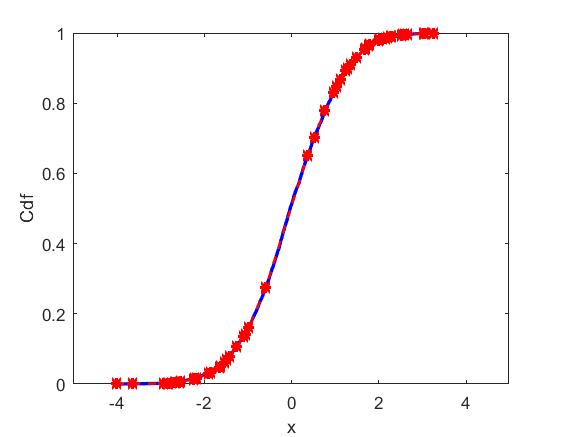}
		\caption{}
		\label{figure:fig12b}
	\end{subfigure}
	\caption{Dataset sampled from a Gaussian distribution. (a) Histogram and UMM pdf (solid line). (b) Points of $S$, ecdf (solid line) and UMM cdf (dashed line).}
	\label{figure:fig12}
\end{figure}

\begin{figure}[H]
\centering	
	\begin{subfigure}[b]{0.4\linewidth}
		\includegraphics[width=\linewidth]{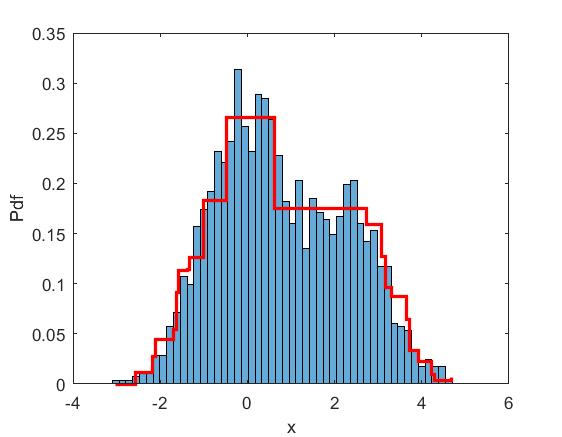}
		\caption{}
		\label{figure:fig13a}
	\end{subfigure}
	\begin{subfigure}[b]{0.4\linewidth}
		\includegraphics[width=\linewidth]{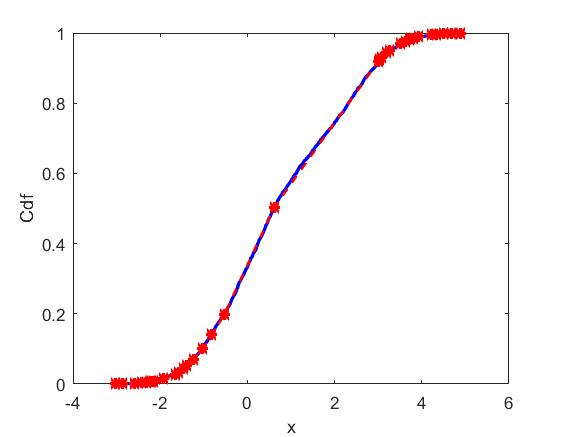}
		\caption{}
		\label{figure:fig13b}
	\end{subfigure}
	\caption{Unimodal dataset sampled from two highly overlapping Gaussians. (a) Histogram and UMM pdf (solid line). (b) Points of $S$, ecdf (solid line) and UMM cdf (dashed line).}
	\label{figure:fig13}
\end{figure}

As already mentioned, in contrast to other unimodality tests, UU-test also achieves to model adequately a unimodal dataset $X$. In unimodal cases, the UU-test directly provides a statistical model, through the final set of $S$ points it returns. The cdf of the statistical model is $PL_S(x)$, which is both unimodal and sufficient approximation of the ecdf. Since the cdf model is piecewise linear, it defines a \textit{Uniform Mixture Model (UMM)} in which each component is the uniform distribution \cite{mclachland,craigmile1997parameter,bouguila2020mixture}. More specifically, if the set provided by UU-test is $S=\{s_1,\ldots, s_{M+1}\}$, then a UMM with $M$ components is defined, where each component $i$ is uniformly distributed in the range $[s_i,s_{i+1}]$, $(i=1,...,M)$. If $N$ is the size of $X$ and $N_i$ is the number of data points in each interval $[s_i,s_{i+1})$, then the UMM pdf is defined as follows:
\begin{displaymath} 
p(x)= \sum_{i=1}^{M} \pi_i \frac{x - s_i}{s_{i+1}-s_i} I(x\in [s_i,s_{i+1})), \quad \pi_i=N_i/N
\end{displaymath}
The corresponding cdf $F(x)$ of the UMM is:
\begin{displaymath} F(x)= \sum_{j=1}^{i-1} \pi_j
+\pi_i \frac{x-s_i}{s_{i+1}-s_i}, \quad s_i\leq x \leq s_{i+1}    
\end{displaymath}
and it is expected to be close to the ecdf. In Figures~\ref{figure:fig10}-\ref{figure:fig13} the UMMs obtained by applying the UU-test on four unimodal datasets are presented both in terms of UMM pdf and of UMM cdf.   
Left subfigures present the histogram and the UMM pdf (solid line), while right subfigures present the points of set $S$, the ecdf (solid line) and the UMM cdf (dashed line).
  
The UMM provided by the UU-test can also be used to \textit{generate synthetic data samples} following the same unimodal distribution as the original dataset using the typical approach for sampling from a mixture model. Fig.~\ref{figure:fig14} refers to a dataset with 2000 points generated by a Gaussian distribution. The histogram and the ecdf of the dataset are presented in Fig.~\ref{figure:fig14a} and Fig.~\ref{figure:fig14b} respectively. The UU-test is applied to this dataset and a UMM model is obtained. Fig.~\ref{figure:fig14c} and Fig.~\ref{figure:fig14d} present the pdf and ecdf of a dataset of the same size that is generated using the UMM model. It is obvious that both histograms and ecdfs are almost identical. 

\vspace{-0.7cm}
\vspace{-0.5cm}

 \begin{figure}[H]
  \centering  
    \begin{subfigure}[b]{0.4\linewidth}
    \includegraphics[width=\linewidth]{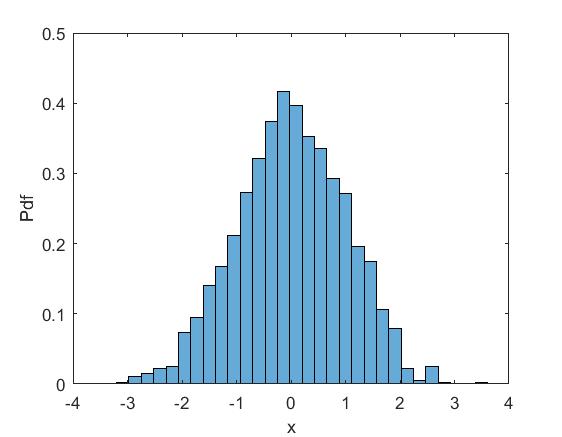}
    \caption{}
    \label{figure:fig14a}
  \end{subfigure}
 \begin{subfigure}[b]{0.4\linewidth}
    \includegraphics[width=\linewidth]{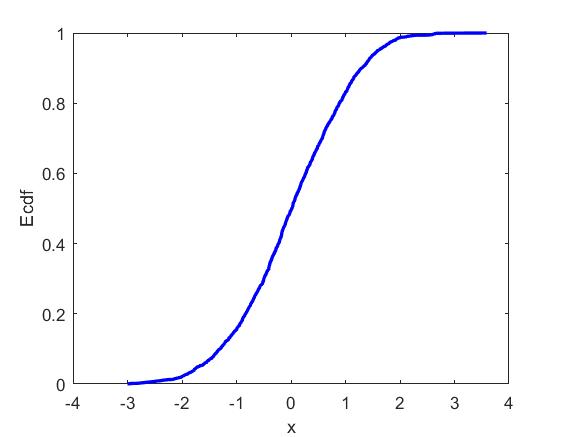}
    \caption{}
    \label{figure:fig14b}
  \end{subfigure}
    \end{figure}
\vspace{-0.5cm}
  \begin{figure}[H]\ContinuedFloat
\centering  
    \begin{subfigure}[b]{0.4\linewidth}
    \includegraphics[width=\linewidth]{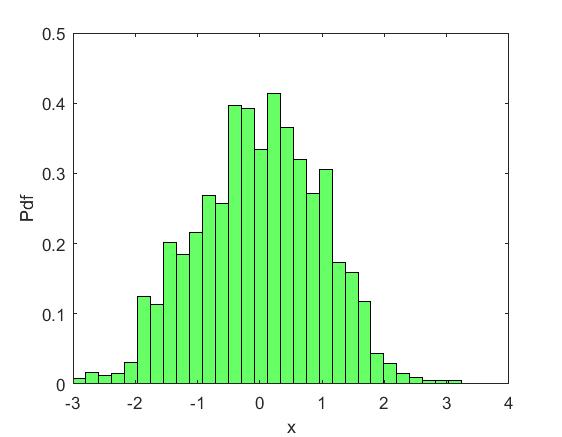}
    \caption{}
    \label{figure:fig14c}
    \end{subfigure}
    \begin{subfigure}[b]{0.4\linewidth}
    \includegraphics[width=\linewidth]{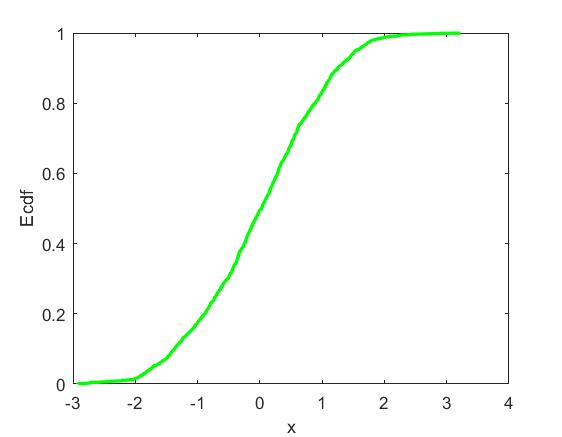}
    \caption{}
    \label{figure:fig14d}
  \end{subfigure}
 \caption{Histogram (a) and ecdf (b) of a dataset $X$ sampled from the Gaussian distribution. Histogram (c) and ecdf (d) of a dataset sampled from the UMM obtained by applying UU-test on the Gaussian dataset $X$ of (a) and (b).}
\label{figure:fig14}
\end{figure}

\section{Experimental results} \label{Experiments}
To assess the effectiveness of the UU-test, we conducted two series of experiments. In the first series, we compared the decisions of UU-test to those of the dip-test using several unimodal and multimodal synthetic and real datasets. In the second series of experiments, our aim was to evaluate the Uniform Mixture Model provided by UU-test as a tool for statistical modeling of unimodal data. 

\subsection{Evaluating UU-test decisions}
This part of our experimental study aims to assess the performance of UU-test in deciding on the unimodality of a dataset. At first, we generated synthetic unimodal and multimodal datasets and computed the decisions of dip-test and UU-test. In addition, we provide two synthetic examples illustrating the influence of noise and outliers on gcm/lcm points and UU-test decision. Finally, we present the results from the application of dip-test and UU-test on the features of several real datasets. 

\subsubsection{Synthetic Datasets}

At this part, we generated datasets from 15 unimodal (U) and multimodal (M) distributions as presented in Table~\ref{table:table2}. The multimodal distributions were mixtures of two or three Gaussians. The parameters and the dataset sizes are shown in the second column of Table~\ref{table:table2}. For each distribution 50 datasets were generated and both UU-test and dip-test were applied on each dataset. Thus, in total 750 synthetic datasets were generated.   

\begin{longtable}{|c|c|c|c|c|} \hline
	Distribution & Parameters&Dip-test (\%)&UU-test (\%)&\makecell{Agreement of \\ two tests (\%)}\\ \hline
	\makecell{Gaussian($\mu,\sigma^2$) (U)}& $\mu=0, ~~ \sigma=1, ~~ N=2000$&$100$&$100$&$100$\\ \hline
	\makecell{Student's t($\nu$)\\ $\nu$: degrees of freedom (U)}&$\nu=4, ~~ N=2000$&$100$&$100$&$100$\\ \hline
	\makecell{Gamma($k,\theta$)\\ $k$: shape, $\theta$: scale  (U)}&$k=1,~~ \theta=2, ~~ N=2000$&$100$&$100$&$100$\\ \hline
	\makecell{Exponential($\lambda$)\\ $\lambda$: rate (U)}&$\lambda=3, ~~ N=2000$&$100$&$100$&$100$\\
	\hline
	\makecell{Cauchy($v$) \\$v$: degrees of freedom (U)}&$v=1, ~~ N=2000$&$100$&$100$&$100$\\ \hline
	\makecell{Triangular ($L,U,m$) (U)\\ $L$: Lower limit, $U$: Upper limit,\\ $m$: mode }&\makecell{$L=-1, ~~ U=1, ~~ m=0$, \vspace{1mm} \\ $N=3700$} &$100$&$100$&$100$\\ \hline
	\makecell{Asymmetric Triangular (U) \\}&\makecell{$L=-4,~~ U=3,~~ m=0$,\vspace{1mm}\\ ~~ $N=6500$}&$100$&$96$&$96$\\ \hline
	\makecell{Two Gaussians  (M)}&\makecell{$\mu_1=0$,~~$\sigma_1=1$, ~~ $N_1=2000$ \vspace{1mm}\\ $\mu_2=4$,~~$\sigma_2=1$, ~~ $N_2=2000$}&$100$&$100$&$100$\\ \hline
	\makecell{Two Gaussians (M)}&\makecell{$\mu_1=0$,~~$\sigma_1=1$, ~~ $N_1=2000$ \vspace{1mm} \\ $\mu_2=4$,~~$\sigma_2=1$, ~~ $N_2=1000$}&$100$&$100$&$100$\\ \hline
	\makecell{Two Gaussians  (U)}&\makecell{$\mu_1=0$,~~$\sigma_1=1$,~~ $N_1=1000$ \vspace{1mm} \\$\mu_2=4$,~~$\sigma_2=2$,~~ $N_2=1000$}&$100$&$100$&$100$\\ \hline
	\makecell{Two Truncated Gaussians (U)\\ with same mean}&\makecell{$\mu_1=0$,~~$\sigma_1=1$,~~ $N_1=1000$ (Left part)\vspace{1mm}\\$\mu_2=0$,~~$\sigma_2=3$,~~ $N_2=1000$ (Right part)}&$100$&$94$&$94$\\ \hline
	\makecell{Three Gaussians  (M)}&\makecell{$\mu_1=0$,~~$\mu_2=4$,~~$\mu_3=8$,\\$\sigma_1=\sigma_2=\sigma_3=1$,\\ $N_1=N_2=N_3=1000$}&$100$&$100$&$100$\\ \hline
	\makecell{Three Gaussians  (M)}&\makecell{$\mu_1=0$,~~$\mu_2=4$,~~$\mu_3=7$,\\$\sigma_1=\sigma_2=\sigma_3=1$,\\ $N_1=N_2=1000,~~ N_3=2000$}&$100$&$100$&$100$\\ \hline
	\makecell{Student's t($\nu$) \& Uniform($a,b$) (U)\\$a$:minimum value\\ $b$:maximum value }& \makecell{$\nu$=10, ~~ $a=0$,~~ $b=10$, ~~ $N=15000$}&$100$&$96$&$96$\\ \hline
	\makecell{Uniform($a,b$) \&  Gaussian($\mu,\sigma^2$) (U)} & \makecell{$a=-10$,~~  $b=5$,~~$\mu=3,~~ \sigma=1$,\vspace{1mm}\\ $N=16000$}&$100$&$96$&$96$ \\ \hline
	\caption{Accuracy of UU-test and dip-test on deciding unimodality (U) or multimodality (M).}
	\label{table:table2}
\end{longtable}

We compared the results of UU-test and dip-test using the same significance level ($\alpha=0.01$). For each distribution, the percentage of 50 datasets for which each test provides correct decision is presented (third and fourth column) as well as the percentage of 50 datasets for which the two tests provided the same decision (fifth column).
It can be observed that UU-test provides in most cases (for 741 out of 750 datasets) correct unimodality decisions that are in agreement with those of the dip-test.

\subsubsection{Examples with noise and outliers}

As it can be expected, noise and outliers affect the existence and position of gcm/lcm points.  
In Fig.~\ref{figure:fig16} we present a bimodal dataset generated from two Gaussians which is distorted by adding uniform noise between the two Gaussians. It can be observed that the  lcm/gcm points (A/B) in the middle of the ecdf (Fig.~\ref{figure:fig16b}) have been eliminated once the noise has been added (Fig.~\ref{figure:fig16d}). Nevertheless, the application of UU-test on the noisy dataset provides the correct decision, i.e. that the dataset remains multimodal.

In Fig.~\ref{figure:fig17} we present a unimodal dataset generated from a single Gaussian, which distorted by the addition of outliers  (left tail) generated from a Student's t distribution. It can be observed that the original gcm points (between A and B) (Fig.~\ref{figure:fig17b}) neither change or move, however, due to the addition of outliers on the left, two new gcm points (C and D) are generated (Fig.~\ref{figure:fig17d}). As with the previous example, the addition of outliers does not modify the result of the UU-test, which decides that the  distorted dataset remains unimodal. 

\begin{figure}[H]
	\centering  
	\begin{subfigure}[b]{0.4\linewidth}
		\includegraphics[width=\linewidth]{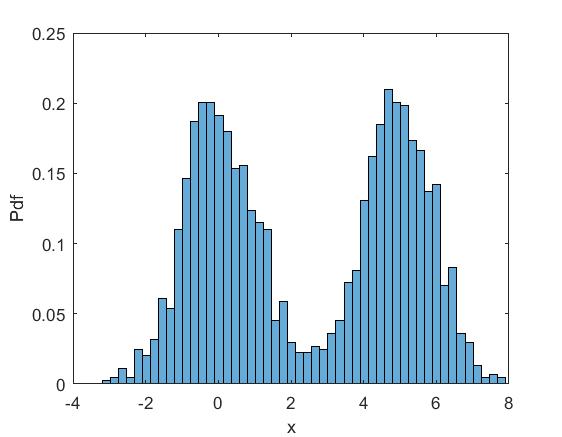}
		\caption{}
		\label{figure:fig16a}
	\end{subfigure}
	\begin{subfigure}[b]{0.4\linewidth}
		\includegraphics[width=\linewidth]{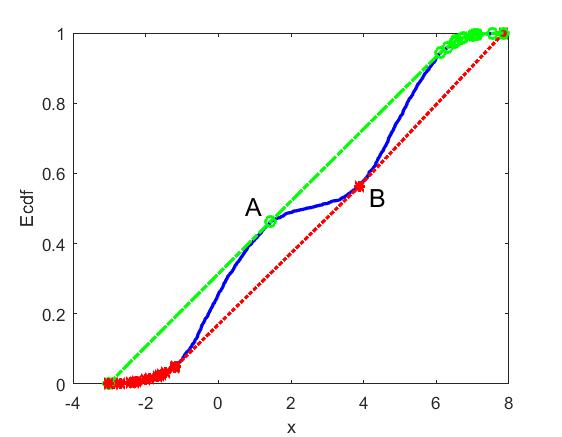}
		\caption{}
		\label{figure:fig16b}
	\end{subfigure}
\end{figure}
\vspace{-0.5cm}
\begin{figure}[H]\ContinuedFloat
	\centering  
	\begin{subfigure}[b]{0.4\linewidth}
		\includegraphics[width=\linewidth]{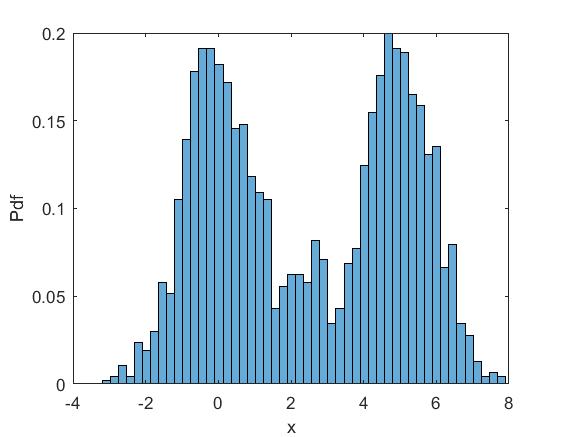}
		\caption{}
		\label{figure:fig16c}
	\end{subfigure}
	\begin{subfigure}[b]{0.4\linewidth}
		\includegraphics[width=\linewidth]{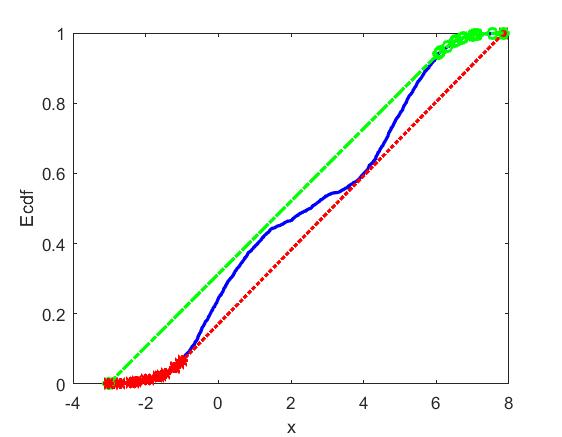}
		\caption{}
		\label{figure:fig16d}
	\end{subfigure}
	\caption{Top row: histogram and ecdf of a bimodal dataset generated by two Gaussians. A and B are middle lcm and gcm points respectively. Bottom row: histogram and ecdf of the dataset after adding uniform noise between the Gaussians. The middle lcm/gcm points A and B have been eliminated, however, UU-test still decides multimodality.}
	\label{figure:fig16}
\end{figure}

It should be noted that, in case we wish to explicitly deal with noise and outliers, we could approximate the ecdf using an appropriate regression method, and then work (e.g. compute gcm and lcm points) with the obtained regression model. Such an approach has been successfully applied in \cite{chen2005seeking} where the image histogram is approximated using support vector regression and the obtained support vectors are exploited to appropriate segmentation thresholds.

\subsubsection{Real Datasets}

We also applied dip-test and UU-test on each feature of five known real datasets, namely Iris, Banknote and Seeds from the UCI Machine Learning Repository \cite{Dua:2019}, Prestige \cite{prestige} and House \cite{kaggle}.
Table \ref{table:table3} presents the datasets and the decision (unimodality (U) or multimodality (M)) of dip-test and UU-test on each dataset feature. Note that the ground truth decision for each feature is not available. The two tests agree on all dataset features except for feature 14 of House dataset. This feature is unimodal based on dip-test and multimodal based on UU-test. Fig.~\ref{figure:fig21} presents the histogram and ecdf of this feature. As it can be observed, this is a borderline case. 

\begin{figure}[H]
	\centering  
	\begin{subfigure}[b]{0.4\linewidth}
		\includegraphics[width=\linewidth]{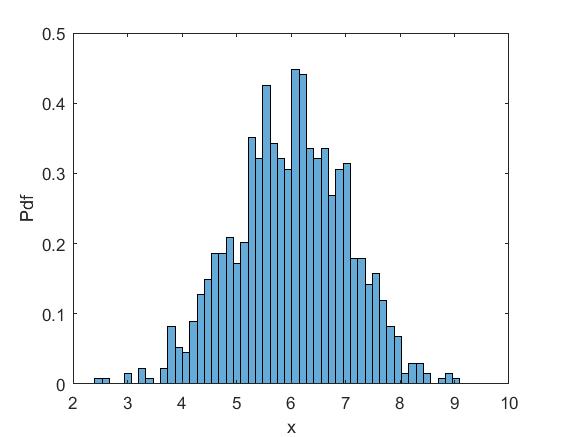}
		\caption{}
		\label{figure:fig17a}
	\end{subfigure}
	\begin{subfigure}[b]{0.4\linewidth}
		\includegraphics[width=\linewidth]{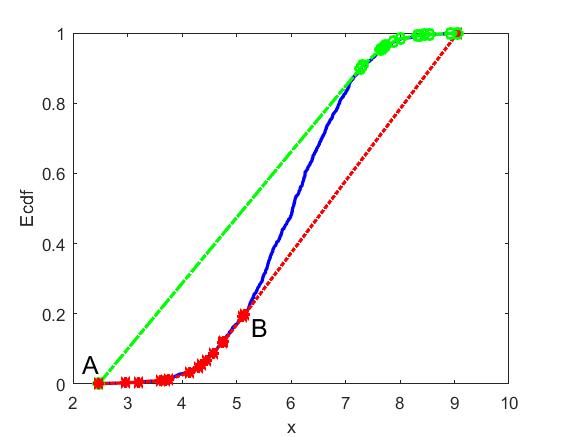}
		\caption{}
		\label{figure:fig17b}
	\end{subfigure}
\end{figure}
\vspace{-0.5cm}
\begin{figure}[H]\ContinuedFloat
	\centering  
	\begin{subfigure}[b]{0.4\linewidth}
		\includegraphics[width=\linewidth]{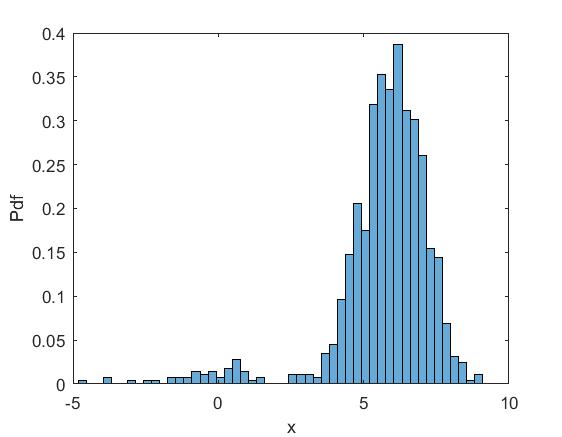}
		\caption{}
		\label{figure:fig17c}
	\end{subfigure}
	\begin{subfigure}[b]{0.4\linewidth}
		\includegraphics[width=\linewidth]{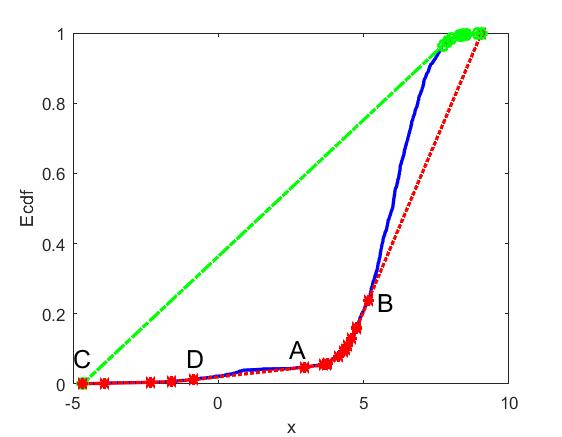}
		\caption{}
		\label{figure:fig17d}
	\end{subfigure}
	\caption{Top row: histogram and ecdf of a dataset generated by a single Gaussian. The gcm points between A and B are illustrated. Bottom row: histogram and ecdf of the dataset after adding Student's t distributed noise (outliers) on the left. Two new gcm points (C and D) have been generated, however, UU-test still decides unimodality.}
	\label{figure:fig17}
\end{figure}

 \begin{figure}[H]
	\centering   
	\begin{subfigure}[b]{0.40\linewidth}
		\includegraphics[width=\linewidth]{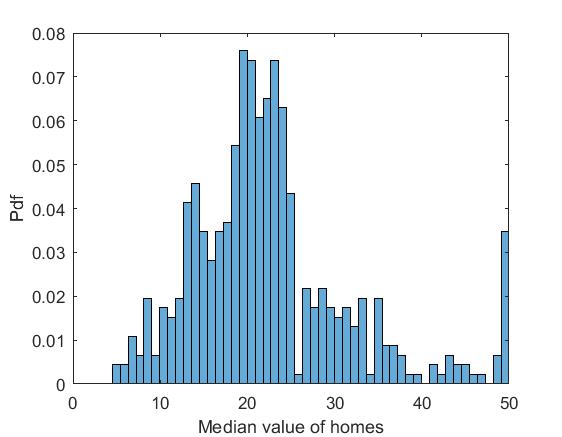}
	\end{subfigure}
	\begin{subfigure}[b]{0.40\linewidth}
		\includegraphics[width=\linewidth]{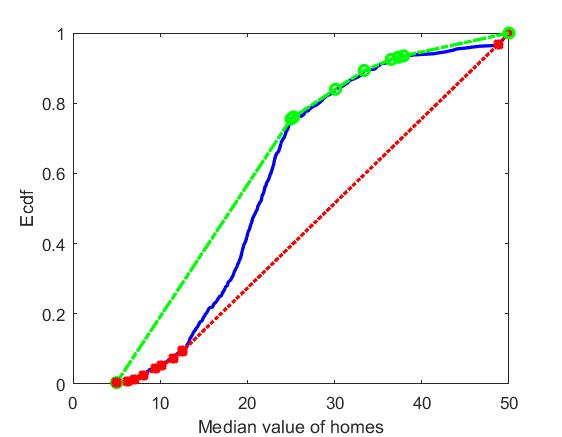}
	\end{subfigure}
	\caption{Histogram and ecdf of feature 14 of House dataset \cite{kaggle} for which dip-test decides unimodality and UU-test decides multimodality.}
	\label{figure:fig21}
\end{figure}

\vspace{1.5cm}

\begin{longtable}{|c|c|c|c|c|} \hline
	Datasets & Features&Dip-test &UU-test &\makecell{Agreement of \\ two tests}\\ \hline
	\multirow{ 2}{*}{Iris}& 1-2&U&U&yes \\ 	\cline{2-5}
	& 3-4&M &M &yes \\ \hline	
	\multirow{ 3}{*}{Banknote}& 1&U&U&yes \\ \cline{2-5}
	& 2&M&M&yes \\ \cline{2-5}	
	& 3-4&U&U&yes \\ \hline	
	Seeds& 1-7&U&U&yes \\ \hline	
	\multirow{ 2}{*}{Prestige}& 1-4&U&U&yes \\ \cline{2-5}	
	& 5&M&M&yes \\ \hline
	\multirow{ 5}{*}{House}& 1&U&U&yes \\ \cline{2-5}
	& 2-5&M&M&yes \\ \cline{2-5}
	& 6-8&U&U&yes \\ \cline{2-5}
	& 9-11&M&M&yes \\ \cline{2-5}
	& 12-13&U&U&yes \\ \cline{2-5}
	& 14&U&M&no \\ \hline	
	\caption{Dip-test and UU-test unimodality (U) or multimodality (M) decisions on features of real datasets.}
	\label{table:table3}
\end{longtable}

\subsection{Uniform Mixture Modeling of Unimodal Data}

We also conducted a series of experiments using synthetic datasets in order to evaluate the UMM provided by the UU-test. For each dataset, we also fitted a Gaussian model as well as a uniform model. In our experiments we considered a variety of unimodal distributions. Table \ref{table:table4} describes the distributions, their parameters and the size of training and test set. In Fig.~\ref{figure:fig15}, we present for each dataset the pdfs of the three fitted models: Gaussian (left figure), Uniform (middle figure) and UMM (right figure). It can be clearly observed that the UMMs provided by the UU-test constitute accurate statistical models of the datasets.
   
In order to measure the quality of the three statistical models, two criteria were considered. The first one is the log-likelihood on a test set and the results are presented in Table~\ref{table:table5}. We used $75\%$ of the sample without replacement as a test set. The rest $25\%$ was used as a training set to build the UMM, Gaussian and Uniform models. Then we computed the log-likelihood of each model on the test set (higher values imply better fit). 
\vspace{3cm}
\begin{longtable}{|c|c|c|c|} \hline
	Distribution & Parameters & Size of training set & Size of test set \\ \hline
	\makecell{Gaussian($\mu,\sigma^2$)}& $\mu=0, ~~ \sigma=1$&$650$&$2000$\\ \hline
	\makecell{Student's t($\nu$)}&$\nu=4$&$650$&$2000$\\ \hline
	\makecell{Gamma($k,\theta$)}&$k=1,~~ \theta=2$&$650$&$2000$\\ \hline
	\makecell{Triangular ($L,U,m$)}&$L=-1,~~ U=1,~~ m=0$ &$12500$&$37000$\\ \hline
	\makecell{Asymmetric Triangular}&$L=-4,~~ U=3,~~ m=0$&$2150$&$6500$\\ \hline
	\makecell{Two Gaussians}&\makecell{$\mu_1=0$,~~$\sigma_1=1$\\ $\mu_2=3$,~~$\sigma_2=1$}&$5850$&$17500$\\ \hline
	\makecell{Student's t($\nu$) \& Uniform($a,b$)}& \makecell{$\nu=10$, ~~ $a=0$,~~ $b=10$} &$5000$&$15000$\\ \hline
	\makecell{Uniform($a,b$) \& Gaussian($\mu,\sigma^2$)} & $a=-10$,~~  $b=5$, ~~$\mu=3,~~ \sigma=1$ &$5300$&$16000$\\ \hline
	\caption{Types and parameters of distributions and size of training and test set of the datasets used for UMM evaluation.}
	\label{table:table4}
\end{longtable}

 \begin{figure}[H]
 \hspace*{-1cm}
    \begin{subfigure}[b]{0.55\linewidth}
    \includegraphics[width=\linewidth]{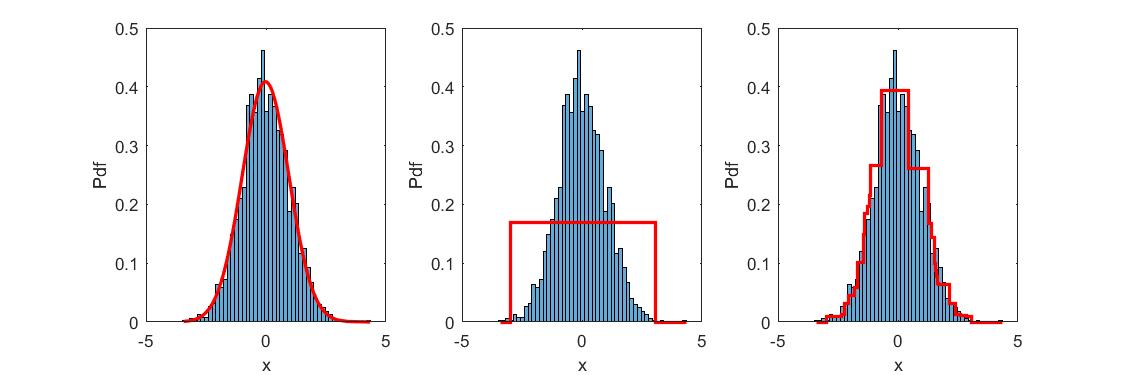}
    \caption{}
    \label{figure:fig15a}
  \end{subfigure}
 \begin{subfigure}[b]{0.55\linewidth}
    \includegraphics[width=\linewidth]{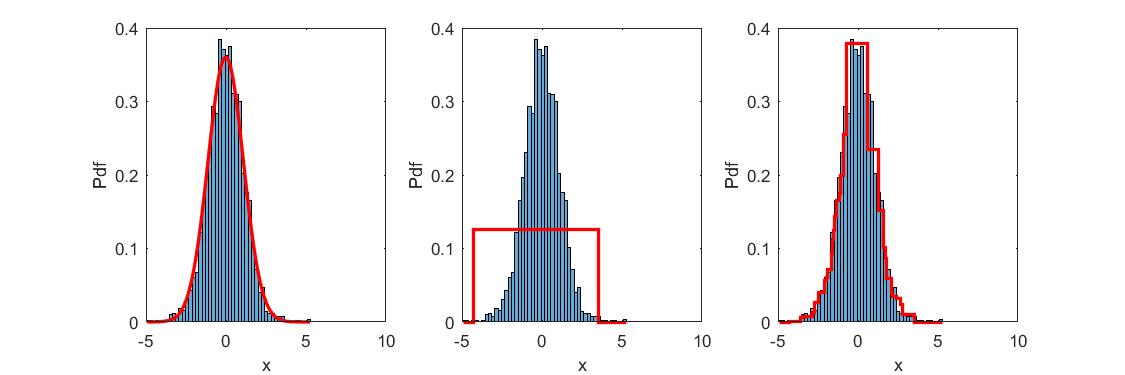}
    \caption{}
    \label{figure:fig15b}
  \end{subfigure}
    \end{figure}
\vspace{-0.5cm}
  \begin{figure}[H]\ContinuedFloat
  \hspace*{-1cm}
    \begin{subfigure}[b]{0.55\linewidth}
    \includegraphics[width=\linewidth]{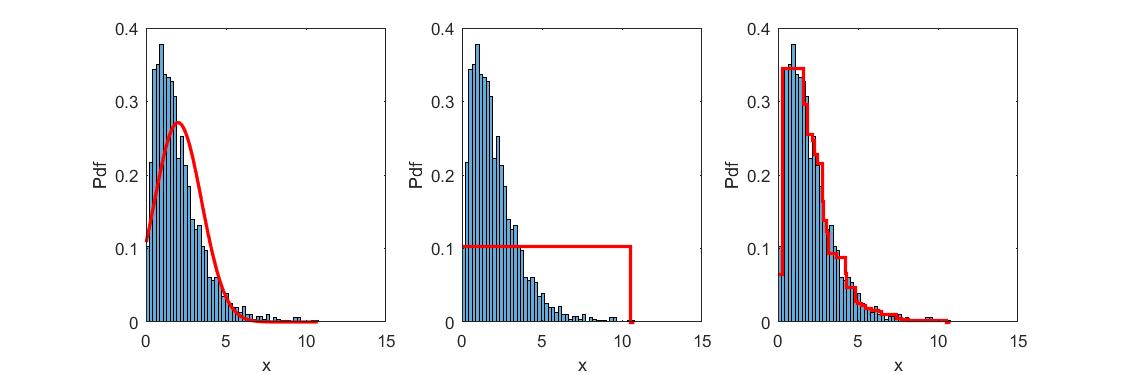}
    \caption{}
    \label{figure:fig15c}
    \end{subfigure}
    \begin{subfigure}[b]{0.55\linewidth}
    \includegraphics[width=\linewidth]{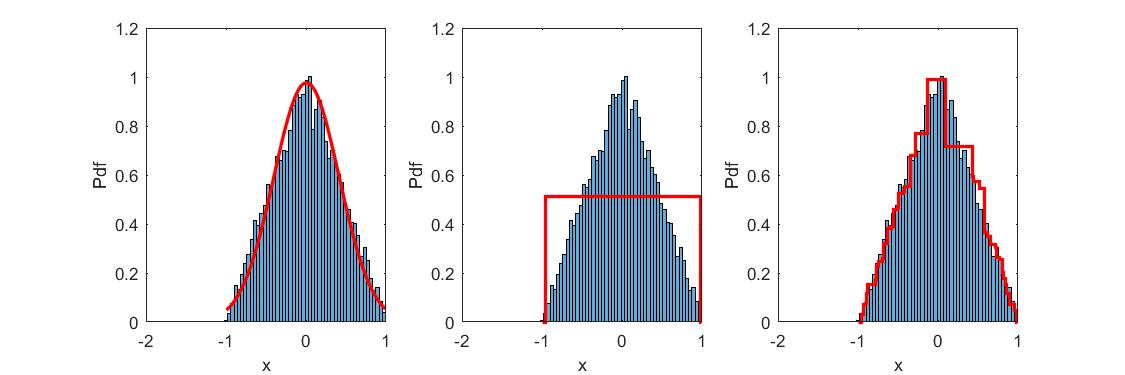}
    \caption{}
    \label{figure:fig15d}
   \end{subfigure}
  \end{figure}
\vspace{-0.5cm}
  \begin{figure}[H]\ContinuedFloat
  \hspace*{-1cm}
  \begin{subfigure}[b]{0.55\linewidth}
    \includegraphics[width=\linewidth]{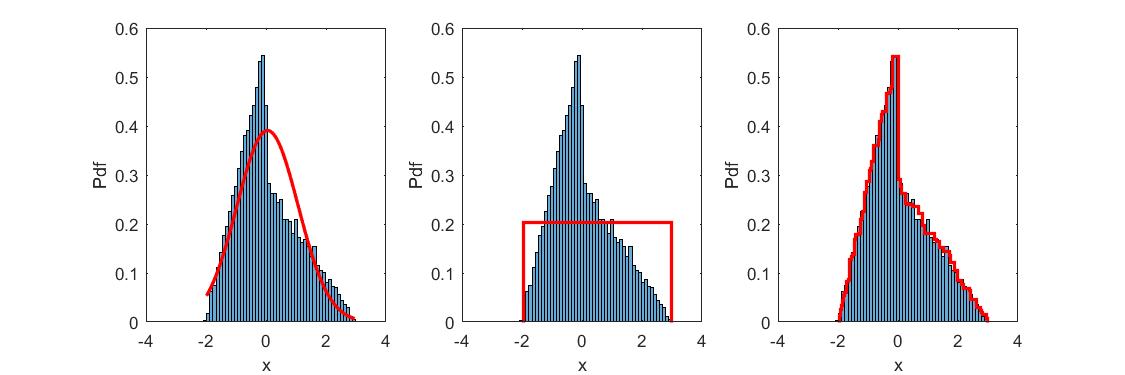}
    \caption{}
    \label{figure:fig15e}
    \end{subfigure}
    \begin{subfigure}[b]{0.55\linewidth}
    \includegraphics[width=\linewidth]{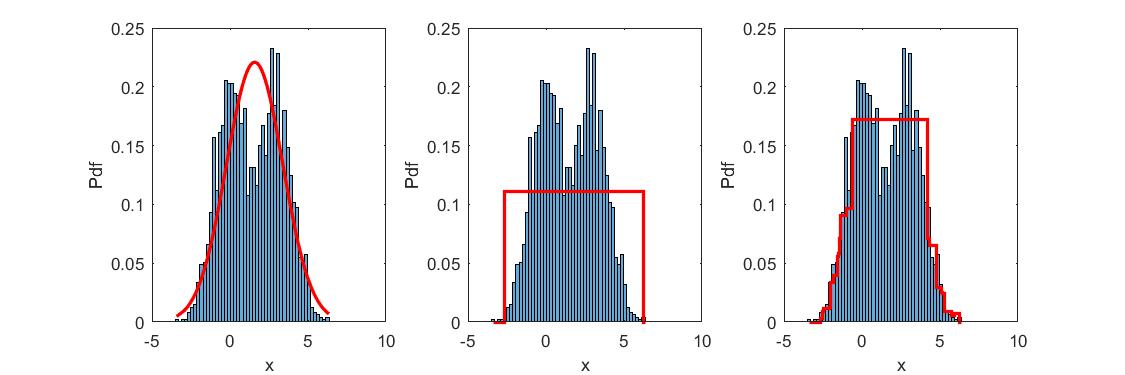}
    \caption{}
    \label{figure:fig15f}
  \end{subfigure}
    \end{figure}
\vspace{-0.5cm}
  \begin{figure}[H]\ContinuedFloat
  \hspace*{-1cm}
  \begin{subfigure}[b]{0.55\linewidth}
    \includegraphics[width=\linewidth]{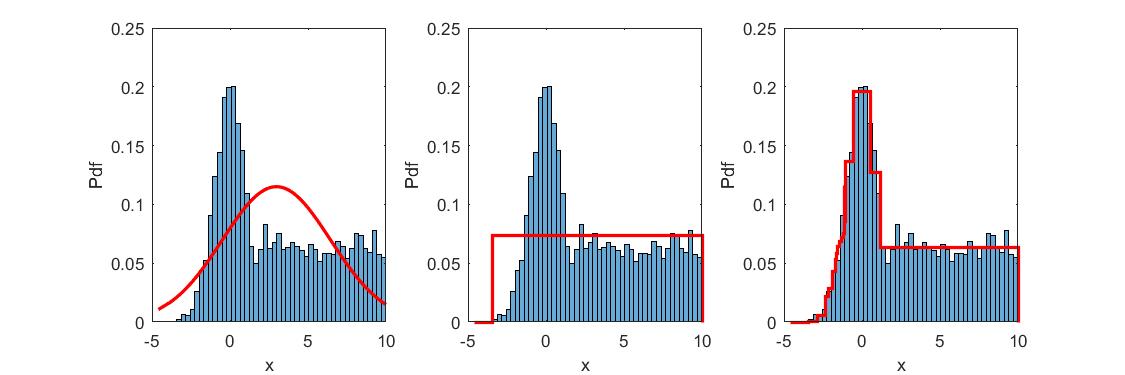}
    \caption{}
    \label{figure:fig15g}
    \end{subfigure}
    \begin{subfigure}[b]{0.55\linewidth}
    \includegraphics[width=\linewidth]{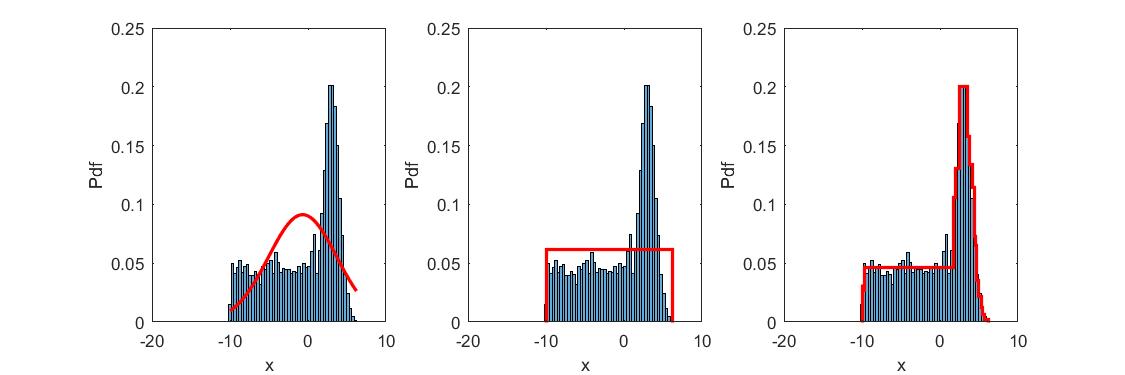}
    \caption{}
    \label{figure:fig15h}
  \end{subfigure}  
\caption{Examples of statistical model fitting on several datasets using Gaussian (left figures), Uniform (middle figures) and UMM (right figures).}
  \label{figure:fig15}
  \end{figure}

\begin{longtable}{|c|c|c|c|} \hline
	Distribution&Gaussian Model&Uniform Model&UMM\\ \hline
	Gaussian&$\mathbf{-13338}$&$-17694$&$-14027$\\ \hline
	Student's t&$-16331$&$-26681$&$\mathbf{-16149}$\\ \hline
	Gamma&$-38283$&$-37044$&$\mathbf{-34591}$\\ \hline
	Triangular&$-19451$&$-25697$&$\mathbf{-18899}$\\ \hline
	Asymmetric Triangular&$-93852$&$-104910$&$\mathbf{-89273}$\\ \hline
	Two Gaussians&$-32877$&$-42027$&$\mathbf{-32483}$\\ \hline
	Student's t \& Uniform&$-40245$&$-43284$&$\mathbf{-36288}$\\ \hline
	Uniform \& Gaussian&$-45959$&$-45828$&$\mathbf{-39153}$\\ \hline
	\caption{Statistical model evaluation using the test set log-likelihood (the higher the better). Bold values indicate the best model in each row.}
	\label{table:table5}
\end{longtable}

In addition, we used the two-sample Kolmogorov-Smirnov test as another criterion to evaluate and compare the three models. The two-sample KS test is a nonparametric hypothesis test that evaluates the difference between the ecdfs of two datasets, by computing the maximum absolute difference between the two ecdfs. Actually the test decides if two datasets have been generated from the same continuous distribution. In each experiment we used a dataset (test set) generated from the ground truth distribution and compared it (using the two-sample KS test) with a dataset generated from each of the three fitted models.  The smaller the distance provided by the KS test, the better the fitted model. The experimental results are provided in Table~\ref{table:table6}.

The experimental results clearly indicate that the UU-test successfully models unimodal data through the UMM it provides. According to the test set likelihood criterion (Table~\ref{table:table5}) the Gaussian model constitutes a better solution only in the case of Gaussian distribution. According to the two-sample KS test criterion, the UMM provides much better results except for the case of Gaussian dataset (Table~\ref{table:table6}). In most cases the difference in performance is notable and becomes much  more higher in the case of asymmetric distributions. 

\vspace{1.8cm}
\begin{longtable}{|c|c|c|c|} \hline
Distribution&Gaussian Model&Uniform Model&UMM\\ \hline
Gaussian&$\mathbf{0.0133}$&$0.1945$&$0.0232$\\ \hline
Student's t&$0.0366$&$0.3451$&$\mathbf{0.0186}$\\ \hline
Gamma&$0.1046$&$0.2350$&$\mathbf{0.0164}$\\ \hline
Triangular&$0.0180$&$0.1260$&$\mathbf{0.0062}$\\ \hline
Asymmetric Triangular&$0.0929$&$0.2072$&$\mathbf{0.005}$\\ \hline
Two Gaussians&$0.0365$&$0.2336$&$\mathbf{0.0055}$\\ \hline
Student's t \& Uniform&$0.1488$&$0.2720$&$\mathbf{0.0065}$\\ \hline
Uniform \& Gaussian&$0.2041$&$0.2703$&$\mathbf{0.0048}$\\ \hline
\caption{Statistical model evaluation using the two-sample KS test (the lower the better). Bold values indicate the best model in each row.}
\label{table:table6}
\end{longtable}

\section{Unimodality in multiple dimensions} \label{Multipledim}

Tackling the unimodality issue for multidimensional datasets is not straightforward. The folding test \cite{siffer2018your} provides a direct approach to assess the 'unimodality character' (or level of unimodality) of a dataset $X$ in multiple dimensions. As  mentioned in Section \ref{Relatedwork}, it is based on the idea of folding up the distribution with respect to a pivot point $s^{\star}$, computing the variance of the folded distribution and finally computing the \textit{folding statistic} ($\Phi(X)$) based on the ratio of the folded variance to the initial variance. Values of $\Phi(X)$ greater or equal to one indicate unimodality of $X$. Although $\Phi(X)$ is easy to compute, the computation of $p$-value relies on bootstraps sampled from the uniform distribution and is computationally heavy especially in multiple dimensions. 

A major concern regarding the folding test is that it relies on the \textit{empirical claim} that folding up a multimodal distribution leads to variance reduction. Therefore, the notion of unimodality is not explicitly involved in folding test computation. It is not difficult to specify distributions where the above claim is not valid, thus folding test fails to provide the correct decision. We next provide two 1-d characteristic examples. According to the folding test, a dataset $X$ sampled from three Gaussians ($\mu_1=-4, \mu_2=0, \mu_3=4$, $\sigma_1=\sigma_2=\sigma_3=0.5$, $N_1=N_2=N_3=2000$ points) is unimodal ($\Phi(X)=1.12$, $p$-value=$0.009$). On the contrary, for this clearly multimodal dataset, dip-test and UU-test agree that it is multimodal. Another example is a dataset generated by a Gaussian ($\mu=0, \sigma=0.5, N_1=2400$ points) and a Uniform ($\alpha=1$, $\beta=4$, $N_2=1600$ points). For this clearly unimodal dataset, the folding test decides multimodality, since $\Phi(X)=0.853$ and $p$-value=$0.01$. On the contrary, dip-test and UU-test correctly decide unimodality.

The most common approach to assess the unimodality character of a multidimensional dataset $X$ is through the exploitation of 1-d unimodality tests. A characteristic example is the {\em dip-dist criterion} which is used in the dip-means clustering algorithm algorithm \cite{kalogeratos2012dip}. The dip-dist criterion decides on the unimodal character of $X$ by exploiting the notion of viewer. A viewer is an arbitrary data point whose role is to suggest on the unimodality of the dataset by forming the set of its distances to all other data points and applying the unimodality test on this set of distances. The idea is that the distribution of the values in this distance vector could reveal information about the cluster structure. In presence of a homogeneous cluster, the distribution of distances is expected to be unimodal. In the case where distinct subclusters exist, the distribution of distances should exhibit distinct modes, with each mode containing the distances to the data objects of each subcluster. Considering each data point as a viewer, the result of unimodality tests on the rows of the distance matrix provide evidence on whether the dataset $X$ contains subclusters or not.

Another way to assess the unimodality character of a multidimensional dataset is based on the assumption that, if a dataset is unimodal, then every 1-d projection of $X$ should be unimodal. To approximately implement this idea the projection  axes should be selected. The skinny-dip method \cite{maurus2016skinny} applies dip-test on the data axes, while the projected dip-means method \cite{chamalis2018projected} applies dip-test both on data axes and PCA axes.

UU-test could directly replace dip-test in the above two approaches. It should be stressed, that UU-test has particular advantages over dip-test. In the case of unimodality, it provides a statistical model in the form of UMM. This can be exploited in the naive Bayes framework \cite{sammut2011encyclopedia,hastie2009elements}: if all features are found unimodal, their joint density can be modeled as a product of UMMs. In another scenario, if the PCA projections \cite{zbMATH01856266,bishop2006pattern} of a multidimensional dataset are unimodal, then each PCA projection can be modeled using a UMM. Since PCA projections are independent, the density of the PCA vector of projections can be modeled as a product of UMMs. 

\subsection{UU-test for clustering}

Another useful property of UU-test (compared to dip-test) is that, in the case of multimodality, it provides information on how to cut (split) the dataset into subsets so as to finally obtain unimodal subsets. This property is particularly useful for designing incremental clustering schemes (based on cluster splitting) \cite{roux2018comparative,boley1998principal,10.5555/2981345.2981381} since it provides information on how to split the multimodal clusters. 

Two illustrative examples are provided next. Fig.~\ref{figure:fig19a} illustrates a 2-d dataset sampled from three Gaussians. It is clear that feature 1 (horizontal axis) is multimodal, while this of feature 2 (vertical axis) is unimodal. Fig.~\ref{figure:fig19b} presents the histogram of the values of multimodal feature 1. We wish to split this set of values and describe how UU-test can be used to determine effective cut points. UU-test fails to accept unimodality, due to the existence of lcm point A before gcm point B in Fig.~\ref{figure:fig19c}. Therefore, it is reasonable to assume that an effective cut point ($cp_1$) exists in the middle between $x_A$ and $x_B$. After splitting the dataset using $cp_1$, we obtain a left subset that is unimodal and a right subset that is bimodal (see Fig.~\ref{figure:fig19d}, Fig.~\ref{figure:fig19e}). Focusing on the right subset, UU-test decides multimodality due to the existence of lcm point C before gcm point D in Fig.~\ref{figure:fig19f}. Therefore, the middle between $x_C$ and $x_D$ specifies a new cutpoint $cp_2$ that further splits the bimodal subset into two unimodal subsets. Fig.~\ref{figure:fig19g}, Fig.~\ref{figure:fig19h} and Fig.~\ref{figure:fig19i} illustrate the final split of the original dataset into three clusters.

Fig.~\ref{figure:fig20} presents another application of the split method on feature 3 of Iris dataset \cite{Dua:2019}. More specifically, we see the histogram and ecdf of the bimodal feature 3. The existence of lcm point A before gcm B indicates multimodality, and the middle between A and B determines and effective cut point.

\begin{figure}[H]
	\centering
	\begin{subfigure}[b]{0.33\linewidth}
		\includegraphics[width=\linewidth]{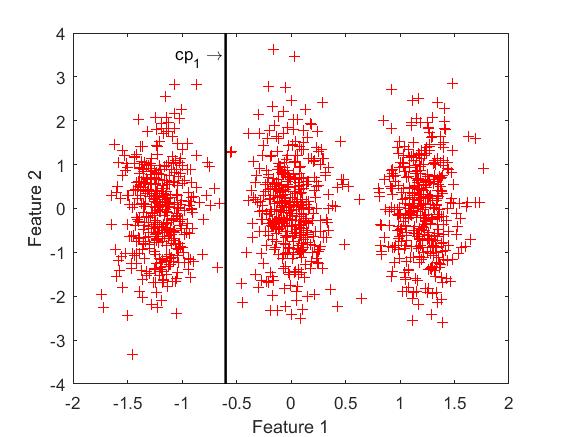}
		\caption{}
		\label{figure:fig19a}
	\end{subfigure}
	\begin{subfigure}[b]{0.33\linewidth}
	\includegraphics[width=\linewidth]{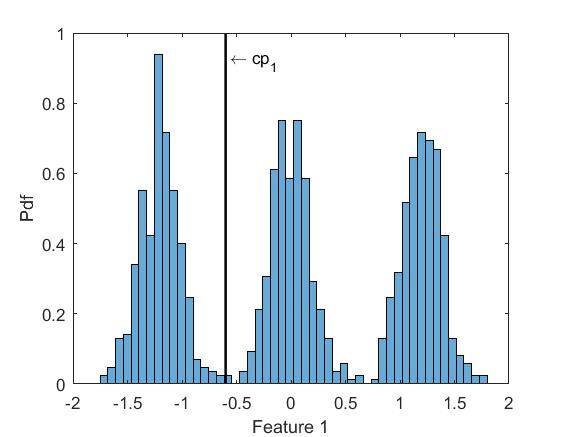}
	\caption{}
	\label{figure:fig19b}
\end{subfigure}
\begin{subfigure}[b]{0.33\linewidth}
	\includegraphics[width=\linewidth]{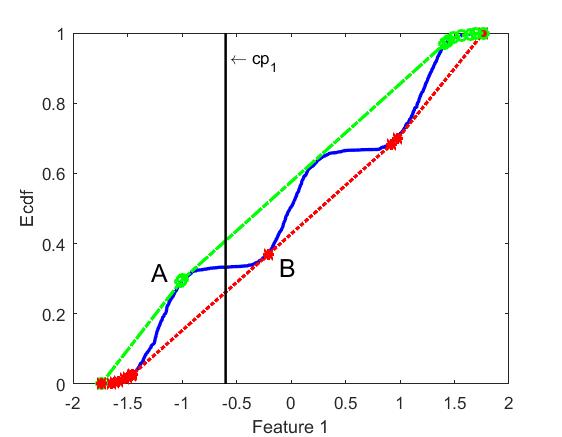}
	\caption{}
	\label{figure:fig19c}
\end{subfigure}
\end{figure}
\vspace{-0.5cm}
\begin{figure}[H]\ContinuedFloat
	\centering  
	\begin{subfigure}[b]{0.33\linewidth}
		\includegraphics[width=\linewidth]{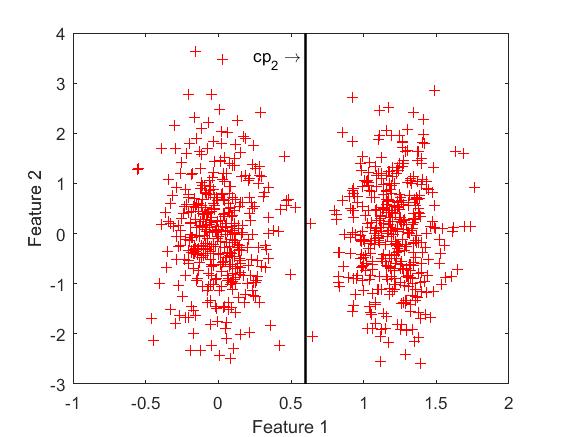}
		\caption{}
		\label{figure:fig19d}
	\end{subfigure}
	\begin{subfigure}[b]{0.33\linewidth}
		\includegraphics[width=\linewidth]{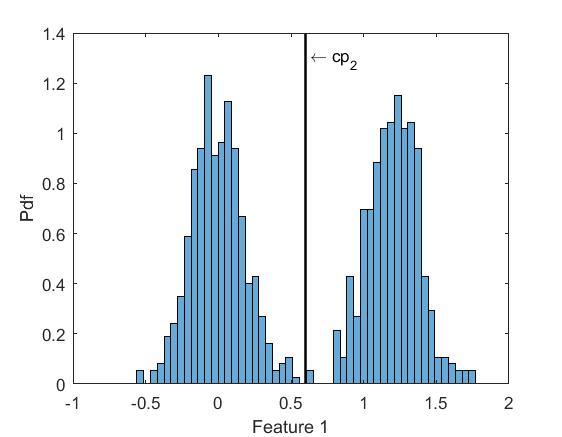}
		\caption{}
		\label{figure:fig19e}
	\end{subfigure}
	\begin{subfigure}[b]{0.33\linewidth}
		\includegraphics[width=\linewidth]{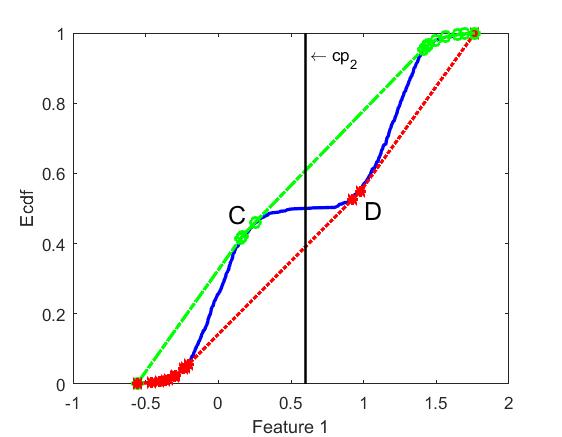}
		\caption{}
		\label{figure:fig19f}
	\end{subfigure}
\end{figure}
\vspace{-0.5cm}
\begin{figure}[H]\ContinuedFloat
	\centering  
	\begin{subfigure}[b]{0.33\linewidth}
		\includegraphics[width=\linewidth]{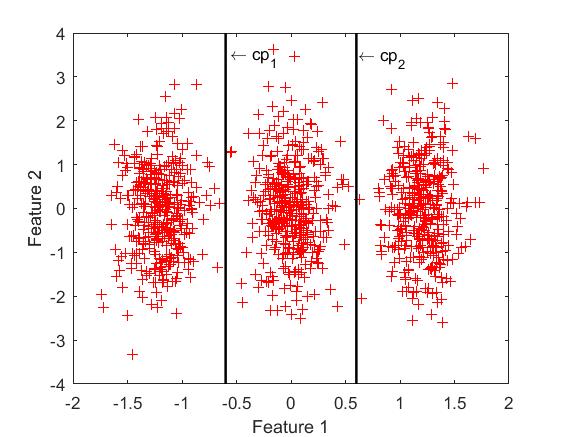}
		\caption{}
		\label{figure:fig19g}
	\end{subfigure}
	\begin{subfigure}[b]{0.33\linewidth}
		\includegraphics[width=\linewidth]{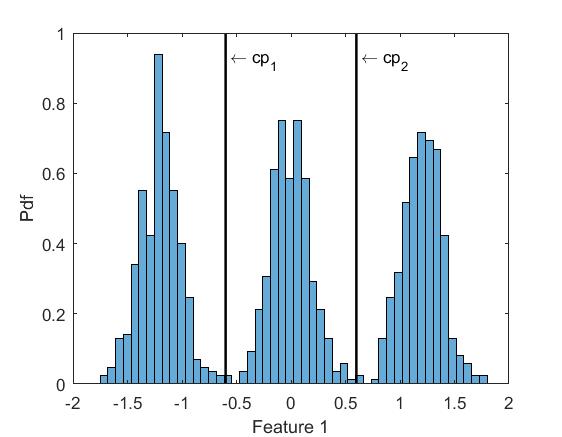}
		\caption{}
		\label{figure:fig19h}
	\end{subfigure}
	\begin{subfigure}[b]{0.33\linewidth}
		\includegraphics[width=\linewidth]{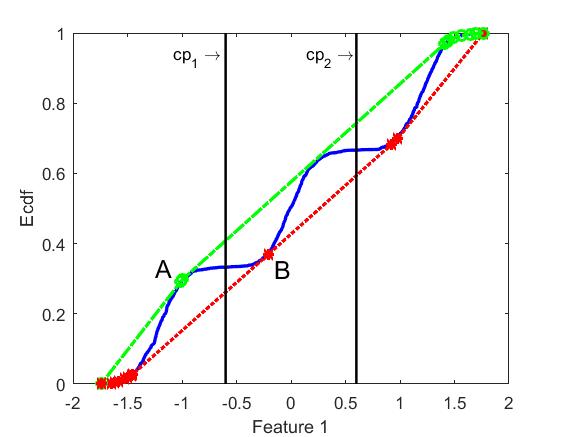}
		\caption{}
		\label{figure:fig19i}
	\end{subfigure}
	\caption{Top row: 2-d plot, histogram and ecdf of feature 1 of a 2-d dataset sampled from three Gaussians. Cut point $cp_1$ is also presented. Middle row: 2-d plot, histogram and ecdf of feature 1 corresponding to the right bimodal subset obtained from the first split. Cut point $cp_2$ is also presented. Bottom row: 2-d plot, histogram and ecdf of feature 1 corresponding to the original dataset along with the two cutpoints.}
	\label{figure:fig19}
\end{figure}
\vspace{-0.6cm}
\begin{figure}[H]
	\centering  
	\begin{subfigure}[b]{0.4\linewidth}
		\includegraphics[width=\linewidth]{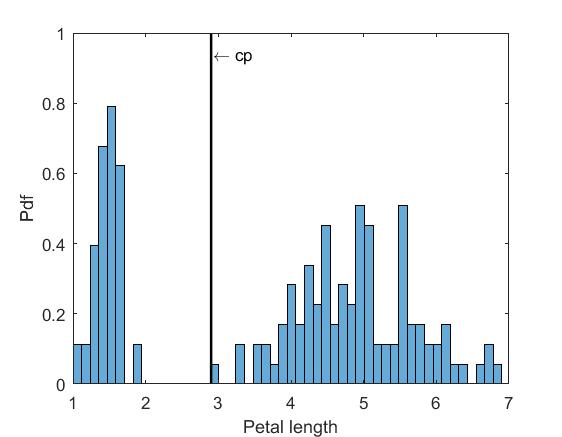}
	\end{subfigure}
	\begin{subfigure}[b]{0.4\linewidth}
		\includegraphics[width=\linewidth]{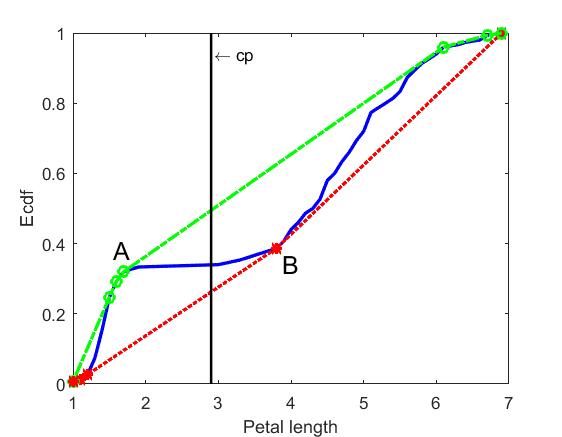}
	\end{subfigure}
	\caption{Histogram and ecdf of feature 3 of Iris dataset \cite{Dua:2019} along with the computed cut point.}
	\label{figure:fig20}
\end{figure}

\section{Conclusions and Future Work} \label{Conclusions}
We have introduced UU-test (Unimodal Uniform test) which is a new method for deciding on dataset unimodality and for statistical modeling of unimodal data. The method takes as input a 1-d dataset and works with the ecdf of the dataset. It attempts to approximate the ecdf by constructing a cdf that is piecewise linear, unimodal and models the data sufficiently. The latter is ensured by applying uniformity (KS) tests on the data subsets corresponding to the linear segments. Unimodality is ensured by first computing the set GL the gcm and lcm points of the ecdf graph and then determining consistent subsets of GL, i.e. subsets where all gcm points lie before the lcm points. In the case where a cdf is found with the above two properties (consistent and sufficient), then UU-test decides unimodality. A unique feature of the method is that it also provides a statistical model of a unimodal dataset in the form of a uniform mixture model (UMM). 

Future research could focus on integration of the UU-test on various data analysis tasks exploiting the decisions on unimodality that it offers. UU-test could be used in clustering algorithms \cite{kalogeratos2012dip,maurus2016skinny} that currently rely on the dip-test for unimodality. As illustrated in Section \ref{Multipledim}, in addition to the decision on unimodality (also provided by dip-test), UU-test directly suggests appropriate cut points in the case of multimodality. Such information is valuable for the clustering algorithm, since the cut points can be used for splitting the multimodal clusters. 

UU-test could also be used in applications that rely on statistical modeling to enhance the typical approach for unimodal data modeling by using the Uniform Mixture Model instead of using a single distribution (e.g. Gaussian, uniform, Student's t etc.). Another line of research concerns the generation of synthetic unimodal data that follow the same distribution as the original unimodal dataset. Moreover, the proposed methodology could be adopted to provide statistical models in the case of multimodal datasets. The main idea is to split the dataset into unimodal subsets and model each unimodal subset using a UMM. Thus, we could obtain a hierarchical statistical model of a multimodal dataset in the form of a mixture of UMMs. 

Finally, the UU-test method could prove useful in image thresholding problems that work with the image histogram \cite{rosin2001unimodal, coudray2010robust, ng2006automatic}. Assessing the unimodal character of the image histogram as well as suggesting appropriate cut points in the case of multimodality, constitute another promising research direction. 

\bibliographystyle{elsarticle-num} 
\bibliography{arxiv-manuscript}
\end{document}